\documentclass[11pt]{article}
\usepackage{fullpage}

\usepackage[utf8]{inputenc} 
\usepackage[T1]{fontenc}    
\usepackage{url}            
\usepackage{array}
\usepackage{booktabs}
\usepackage{tabularx}
\usepackage{amsfonts}       
\usepackage{nicefrac}       
\usepackage{multirow}
\usepackage{makecell}
\usepackage{hhline}
\usepackage{natbib}
\usepackage{mathtools}

\usepackage{tocloft}
\usepackage{braket}
\usepackage[a4paper,margin=1in]{geometry}
\usepackage[toc,page,header]{appendix}
\usepackage{minitoc}
\usepackage{colortbl}

\usepackage[colorlinks,citecolor=blue,urlcolor=blue,linkcolor=blue,linktocpage=true,pagebackref=true]{hyperref}
\renewcommand*{\backref}[1]{}
\renewcommand*{\backrefalt}[4]{(\small%
    \ifcase #1 not cited%
          \or cited on page~#2%
          \else cited on pages #2%
    \fi%
    )}
\usepackage{wustyle}

\usepackage{enumitem}
\usepackage{needspace}

\usepackage{wrapfig}

\usepackage{framed}
\usepackage{placeins}
\usepackage{float}
\usepackage{flafter}

\colorlet{shadecolor}{orange!15}

\definecolor{c4}{RGB}{255,225,187}
\definecolor{c2}{RGB}{209, 233, 184}
\definecolor{c3}{RGB}{218,243,246}
\definecolor{c1}{RGB}{249, 229, 229}
\definecolor{c5}{RGB}{255, 128, 128}
\definecolor{c6}{RGB}{251, 132, 002}

\newcommand{\collapse}{\mathrm{coll}}

\title{\vspace{-2em}
Effective Learning Rate Governs Loss Dynamics \\[-0.15em] in Language Model Pretraining
}

\author{
    Zihan Liu$^{1,2,}$\thanks{Equal contribution; the order  was determined by a fair coin toss.}\quad 
    Ruiheng Zheng$^{1,}$\footnotemark[1]\quad
    Shaobo Zhang$^{1}$\quad
     Changxin Tian$^{2}$\\ Kunlong Chen$^{2}$\quad Zhiqiang Zhang$^{2}$\quad
    Lei Wu$^{1,}$\thanks{Corresponding author: \texttt{leiwu@math.pku.edu.cn}}\\[0.5em]
  $^1$Peking University~~~$^2$Ant Group
}

\date{\vspace{-1em}}

\begin{document}

\maketitle

\doparttoc
\faketableofcontents

\vspace{-1em}
\begin{abstract}
We uncover \textbf{ELR collapse} in language model pretraining: learning rate (LR) and parameter norm govern loss dynamics primarily through their ratio, the effective learning rate (ELR). When ELR is matched across runs, their loss trajectories collapse throughout training despite substantially different LRs and parameter norms.  Across optimizers, architectures, datasets, and model scales, mean collapse errors are typically a few \(\times 10^{-3}\), below the seed-to-seed  variation measured in a representative configuration. Systematic ablations identify normalization design and the timescale of LR--norm variation as key determinants of collapse precision. 
Controlled interventions further show that  weight decay and Hyperball shape loss dynamics primarily through the ELR schedules they induce. Replacing LR with ELR enables a fitted functional scaling law (FSL)~\citep{li2025fsl} to transfer across norm-control methods. The resulting ELR-based FSL also explains delayed acceleration, a recurring effect of norm control.   Together, these results establish ELR as a common coordinate linking LR scheduling, norm control, and loss dynamics.
\end{abstract}

\begin{figure}[H]
    \centering
    \includegraphics[width=0.9\textwidth]{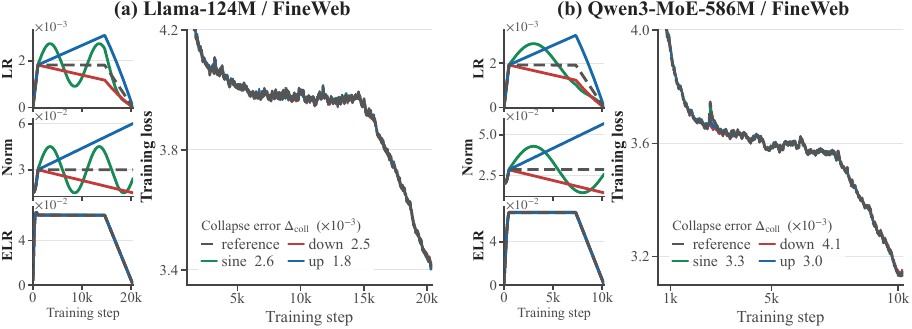}
    \vspace{-.2em}
\caption{
\textbf{ELR matching collapses loss trajectories despite different LR and norm schedules.}
In each panel, three nonconstant LR schedules are paired with prescribed norm schedules to match the ELR of a reference run. Legend values report the post-warmup mean absolute loss discrepancy from the reference in units of \(10^{-3}\). Loss panels are clipped above 4.2 in {\bf (a)} and 4.0 in {\bf (b)} for display.
See Section~\ref{sec:elr_state_variable} and Appendices~\ref{appendix:exp_datails} and~\ref{appendix:collapse_exp} for experimental details and additional results.
}
    \label{fig:elr_collapse}
\end{figure}

\section{Introduction}


\vspace{-.5em}

Training dynamics are primarily governed by  hyperparameters such as the learning rate (LR) and batch size. 
In large language model (LLM) pretraining, norm control provides another axis, encompassing mechanisms that shape parameter-norm evolution either implicitly through weight decay~\citep{krogh1991simple,loshchilov2019adamw} or explicitly through norm constraints~\citep{wen2026hyperball,loshchilov2023weightnormcontrol,xie2026controlledllm}.
It can improve training stability~\citep{dangelo2024needweightdecaymodern} and hyperparameter transfer across  scales~\citep{kosson2026weight}, but it also reshapes loss dynamics and often accelerates convergence. This raises a basic question:
\begin{center}
{\it Does norm control introduce an independent degree of freedom governing loss dynamics?
}
\end{center}

\vspace{-1em}
\paragraph{ELR collapse: a precise macroscopic law.}
Within the regimes studied here, our experiments answer this question in the negative. We construct runs whose LR and parameter-norm trajectories differ substantially while matching a common ELR schedule step by step, where
$
    \eta_k^{\mathrm{eff}}
    \mathrel{:=}
   \eta_k/\|\bW_k\|_F.
$
Figure~\ref{fig:elr_collapse} presents the central experiment:
\begin{center}
\emph{When ELR schedules are matched, full loss trajectories collapse with mean\\ absolute discrepancies of only a few \(\times 10^{-3}\).}
\end{center}
For comparison, changing only the initialization or data-order seed in a representative configuration yields mean pairwise loss discrepancies of order \(10^{-2}\); see Appendix~\ref{appendix:stochastic_baseline}. We call this phenomenon \textbf{ELR collapse}. It suggests a simple macroscopic picture: LR scheduling and norm control shape loss dynamics primarily through the ELR schedule they jointly induce, rather than acting as independent controls.

\vspace{-.5em}
\paragraph{ELR collapse is nontrivial and conditional.}
The precision of ELR collapse does not follow from scale invariance. Transformers are not exactly scale invariant. In a pre-norm residual block,
\vspace{-.5em}
\[
\bh^{\ell+1} = \bh^\ell + \operatorname{FFN}\!\left(\operatorname{Norm}(\bh^\ell)\right),
\]
rescaling \(\bh^\ell\) affects the identity and residual branches differently. Learnable normalization gains, nonhomogeneous activations, and the embedding and output layers introduce further scale dependence. Consequently, ELR collapse is not implied by the exact reparameterization symmetry that motivates ELR in scale-invariant models~\citep{vanlaarhoven2017l2regularizationversusbatch,li2020reconciling}; see Appendix~\ref{appendix:scale-invariance} for a detailed discussion.

High-precision collapse is also not automatic. Our ablations identify three factors that strongly affect its precision: QK-Norm~\citep{henry2020querykey}, learnable RMSNorm gains, and sufficiently gradual variation in the LR--norm realization of a prescribed ELR schedule. Removing QK-Norm weakens the collapse even though training remains stable. Fixing the RMSNorm gains weakens it further, despite making the parameterization more scale invariant. These findings rule out static scale symmetry as a complete explanation and delineate an empirical regime for high-precision collapse, while its microscopic dynamical origin remains open.

\vspace{-.5em}
\paragraph{Beyond collapse: mediation and transfer.}
The prescribed-ELR experiments establish the central phenomenon: different LR and norm realizations of the same ELR schedule produce nearly identical loss trajectories. We next subject the claim that ELR governs loss dynamics to two stronger tests.

\vspace{-.5em}
\begin{itemize}
  \item \textbf{From prescribed norms to practical norm control.}
The prescribed-ELR experiments establish collapse by directly controlling the parameter norm trajectory. We next ask whether the same principle holds under practical norm-control mechanisms. For weight decay and Hyperball~\citep{wen2026hyperball}, we leave the norm control in place and adapt only the LR of a comparison run to match the target ELR schedule. This recovers the target loss dynamics with collapse errors of \(4.8\times10^{-3}\) and \(1.2\times10^{-3}\), respectively. Thus, ELR collapse extends beyond prescribed norm trajectories to practical norm-control methods, whose effects on loss are largely captured by the ELR schedule they induce.

\item \textbf{From trajectory matching to scaling-law transfer.}
We next ask whether ELR enables a predictive scaling law to transfer across norm-control methods. We fit the functional scaling law (FSL)~\citep{li2025fsl} using only non-Hyperball runs and evaluate it, without refitting, on Hyperball runs absent from the fitting set. When parameterized by ELR, the fitted FSL remains accurate; when parameterized by LR, it develops a large systematic bias, increasing the prediction error by \(12.9\times\). Thus, ELR does more than align individual trajectories: it enables FSL to transfer across distinct norm-control methods.
\end{itemize}

\vspace{-1.5em}
\paragraph{ELR explains delayed acceleration.}
Finally, we use the ELR picture to explain the effect of \emph{delayed acceleration}: a norm-controlled run can initially have higher loss, yet later overtake an uncontrolled baseline~\citep{hoffmann2022training,liu2025muon,wen2026hyperball}. We show that this reversal is captured by the ELR schedule induced by norm control and can be reproduced through direct norm control. Thus, delayed acceleration emerges as a consequence of ELR reshaping rather than a mechanism specific to weight decay or Hyperball.

Taken together, our results establish ELR collapse as a precise, conditional, and operational macroscopic law for pretraining loss dynamics. At the level of training loss, LR scheduling and norm control are distinct mechanisms for realizing a common ELR schedule, rather than independent dynamical coordinates.

\vspace{-.4em}
\section{Related Work}
\vspace{-.3em}

\paragraph{ELR under scale invariance.}
Normalization has motivated extensive study of scale-invariant objectives satisfying $\cL(c\bW)= \cL(\bW)$ for all $c>0$. Such objectives depend on $\bW$ only through its direction $\widehat{\bW}=\bW/\|\bW\|$. For gradient descent, the ELR governing the dynamics of \(\widehat{\bW}_k\) is \(\eta_k/\|\bW_k\|^2\)~\citep{hoffer2018normmattersefficientaccurate}. For scale-insensitive optimizers such as signSGD and Adam, the corresponding ELR is \(\eta_k/\|\bW_k\|\)~\citep{vanlaarhoven2017l2regularizationversusbatch}. ELR has since been widely used to analyze the optimization dynamics of normalized models~\citep{li2019exponentiallearningrateschedule,heo2020adamp,li2020reconciling,wan2021sphericalmotiondynamicslearning,kodryan2022threeregimes}.
This line of work derives ELR from exact scale invariance. In contrast, we show that ELR governs loss dynamics to high precision in language model pretraining despite the lack of such symmetry.
\vspace{-.5em}
\paragraph{Norm control in LLM pretraining.}
Weight decay is increasingly understood as a dynamical mechanism that regulates  norms and thereby controls effective angular updates for scale-invariant models, rather than solely as a  regularizer~\citep{zhang2019mechanismsweightdecayregularization,wan2021sphericalmotiondynamicslearning,kosson2024rotationalequilibriumweightdecay}. In  LLM pretraining,  weight decay has been found important for training stability and hyperparameter transfer across scales~\citep{dangelo2024needweightdecaymodern,wortsman2024small,kosson2026weight}. \citet{loshchilov2023weightnormcontrol} proposed weight norm control, which generalizes decoupled weight decay by steering parameter norms toward prescribed targets. Recent methods impose explicit norm constraints: Hyperball constrains parameters to prescribed Frobenius-norm spheres~\citep{wen2026hyperball}, whereas SSO imposes spectral-norm constraints~\citep{xie2026controlledllm}.

Prior experiments also provide  evidence that norm control affects loss through ELR.  \citet{dangelo2024needweightdecaymodern} adjusted the LR of runs without weight decay to reproduce the ELR induced by weight decay, obtaining approximate loss alignment. \citet{li2025efficienthyperparametertuningtrajectory} later reported late-stage loss alignment when \(\eta\lambda\), an equilibrium proxy for ELR, is matched. We extend these observations into a quantitative empirical law: ELR matching yields collapse errors of order \(10^{-3}\) across diverse training configurations, and we identify the conditions required for collapse at this precision. We further test whether the effects of both weight decay and explicit norm control on loss dynamics are indeed governed by ELR.

\vspace{-.5em}
\paragraph{Functional scaling laws.}
\citet{li2025fsl} showed that functional scaling laws (FSL) provide a reliable model for predicting loss dynamics in LLM pretraining under general LR schedules. We show that replacing LR with ELR improves predictive accuracy and, more importantly, enables the fitted FSL to transfer across norm-control regimes. The resulting elr-FSL explains why norm-controlled runs can initially have higher loss than the uncontrolled baselines, yet later overtake them and attain lower final loss---a phenomenon we call \emph{delayed acceleration}.

\vspace{-.5em}

\paragraph{Concurrent work.}
Two concurrent studies examine norm-aware effective step sizes in LLM pretraining. \citet{yang2026nonlinearity} show that expressing the optimal LR in terms of ELR reduces its nonlinear scaling and improves extrapolation across model sizes and data budgets. \citet{xiao2026hyperballmaynot} derive an angular ELR and match it between weight-decayed and Hyperball-controlled Muon runs, approximately recovering the corresponding loss dynamics. Our work addresses the broader trajectory-level hypothesis that ELR governs loss dynamics across training configurations. We test this hypothesis quantitatively and systematically: across optimizers, architectures, datasets, and model scales, matching ELR aligns full loss trajectories with mean discrepancies of only a few \(\times 10^{-3}\). We further identify the conditions controlling this precision and show that ELR enables FSL to transfer across norm-control regimes.


\section{ELR, Norm Control, and Collapse Metrics}
\label{sec:preliminaries}


\paragraph{Effective learning rate and norm control.}
Let \(\bW_k\) be a matrix-valued parameter in a Transformer updated with LR \(\eta_k\) at step $k$. We define its effective learning rate (ELR) as
\begin{align}
    \eta_k^{\eff}
    \coloneqq
    \frac{\eta_k}{\|\bW_k\|_F},
    \label{eq:elr}
\end{align}
and refer to \(\{\eta_k^{\eff}\}_k\) as its ELR schedule. Decoupled weight decay updates \(\bW_k\) according to
\[
    \bW_{k+1}
    =
    (1-\eta_k\lambda)\bW_k-\eta_k\bU_k,
\]
where \(\bU_k\) denotes the optimizer update. Rather than prescribing \(\|\bW_k\|_F\), weight decay regulates its evolution through the shrinkage factor \(1-\eta_k\lambda\), thereby shaping ELR implicitly. Hyperball~\citep{wen2026hyperball}, by contrast, constrains each matrix parameter to a sphere of prescribed radius $R$:
\[
    \bW_{k+1}
    =
    \texttt{Norm}_{R}\!\left(
        \bW_k-\eta_k\,\texttt{Norm}_{R}(\bU_k)
    \right),
    \qquad
    \texttt{Norm}_{R}(\bQ)
    \coloneqq
    R\,\frac{\bQ}{\|\bQ\|_F}.
\]
 Note that the coefficient multiplying $\bU_k$ before the outer projection is  $\tilde{\eta}_k = \eta_kR/\|\bU_k\|_F$. Since the weight norm is $R$, the corresponding ELR is $\eta^{\eff}_k = \tilde{\eta}_k/\|\bW_k\|_F = \eta_k/\|\bU_k\|_F$ for Hyperball.

\paragraph{Quantifying ELR collapse.}
Each collapse experiment is a paired controlled comparison. The target and
comparison runs use the same parameter initialization and data order, with all
other sources of training randomness held fixed. They differ only in the LR
and/or norm-control intervention under study, resulting in different LR and
parameter-norm trajectories while their ELR trajectories are matched.
Let $\{L_k\}$ and $\{L_k^{\mathrm{ref}}\}$ denote the loss trajectories of a matched run and its reference run, respectively. We define the loss residual and mean collapse error as
\begin{equation}
    r_k
    \coloneqq
    L_k-L_k^{\mathrm{ref}},
    \qquad
    \Delta_{\collapse}
    \coloneqq
    \frac{1}{|\mathcal T|}
    \sum_{k\in\mathcal T}|r_k|,
\end{equation}
where \(\mathcal T\) denotes the set of training steps over which collapse is evaluated.

\paragraph{Calibrating the scale of collapse error.}
There is no configuration-independent threshold for when two pretraining loss
trajectories differ meaningfully. Nevertheless, recent LLM optimizer studies
routinely use \(10^{-2}\)-scale loss differences to distinguish optimizer and
hyperparameter choices
\citep{liu2024sophia,liu2025muon,wang2025gradpower,wang2025sharpness,chen2026cautious}.
This is a practical convention rather than a universal threshold.

For an internal calibration, we measure run-to-run variation in our most extensively tested configuration, Llama-124M trained on FineWeb. Repeating the same configuration while changing only the initialization seed or data-order seed yields variations of \(1.11\times10^{-2}\) and \(1.62\times10^{-2}\), respectively, under the same metric; see Appendix~\ref{appendix:stochastic_baseline}. Both substantially exceed the collapse errors of a few \(\times10^{-3}\) observed in this setting.

Together, these references place the typical collapse error of a few
\(\times10^{-3}\) well below both seed-induced variation in our reference
configuration and the \(10^{-2}\)-scale improvements regarded as meaningful
in LLM optimizer studies.

\section{The  ELR Collapse}
\label{sec:elr_state_variable}

\textbf{Matching ELR.}
Consider a reference run \(A\) with LR schedule \(\{\eta_k^A\}_{k}\) and norm evolution \(R_k^A\coloneqq\|\bW_k^A\|_F\), which induce the ELR schedule \(\gamma_k\coloneqq\eta_k^A/R_k^A\). For a second run \(B\) with a different LR schedule \(\{\eta_k^B\}_{k}\), matching \(\gamma_k\) requires the target norm \(R_k^B\coloneqq\eta_k^B/\gamma_k\). We enforce \(\|\bW_k^B\|_F=R_k^B\) at every optimization step. By construction, runs \(A\) and \(B\) have different LR and norm schedules but the same ELR schedule. More details are deferred to Appendix~\ref{sub:elr_matching_protocols}.

\subsection{Quantitative Validation across Training Configurations}
\label{sec:quantitative_validation}

We begin with two representative systems: a dense Llama model with 124M
parameters and a Qwen3-MoE model with 586M parameters, both trained with
AdamW. In each setting, we prescribe a common warmup--stable--decay (WSD) ELR
schedule and realize it through four different LR and target-norm schedules.
The reference run uses the WSD LR schedule, while the other three use
linear-up, linear-down, and sinusoidal LRs, with their target norms adjusted
to preserve the prescribed ELR.

\begin{figure}[!h]
    \centering

    \hspace{-1em}\includegraphics[width=0.91\textwidth]{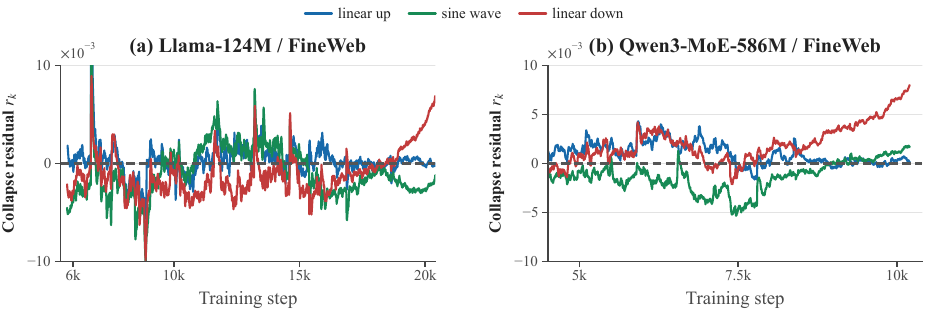}
    \vspace{-.5em}
\caption{\textbf{ELR-matched residuals remain centered near zero without systematic drift.}
Signed loss residuals \(r_k=L_k-L_k^{\mathrm{ref}}\) for the linear-up, linear-down,
and sinusoidal LR schedules over the post-warmup evaluation window. Both
panels use the prescribed-ELR protocol of
Figure~\ref{fig:elr_collapse}; the dashed line marks zero residual.}
    \label{fig: collapse-error-dynamics}
    \vspace{-1em}
\end{figure}

Figure~\ref{fig:elr_collapse} presents the central result. Despite substantial
differences in the LR and parameter-norm trajectories, the corresponding loss
curves nearly coincide throughout training. In the order linear-up,
linear-down, and sinusoidal, the mean collapse errors are
\[
\begin{aligned}
    \text{Llama (124M):}\qquad
    &\Delta_{\mathrm{coll}}
    =(1.8,\,2.5,\,2.6)\times10^{-3},\\
    \text{Qwen3-MoE (586M):}\qquad
    &\Delta_{\mathrm{coll}}
    =(3.0,\,4.1,\,3.3)\times10^{-3}.
\end{aligned}
\]
Figure~\ref{fig: collapse-error-dynamics} further shows the corresponding residuals
\(r_k\) throughout training. They generally fluctuate around zero without
systematic drift, although transient deviations occur. ELR collapse is
therefore a trajectory-level quantitative agreement with mean discrepancies of
only a few \(\times10^{-3}\), rather than exact pointwise equality of the loss
curves.

We next test the same protocol across a broader range of configurations:

\begin{itemize}[leftmargin=2.5em, topsep=2pt, itemsep=2pt, parsep=2pt]
\item \textbf{Models:} Transformer architectures spanning dense (Llama and Qwen3), MoE (Qwen3-MoE), and linear attention based on Kimi Delta Attention (KDA)~\citep{kimi2025kimilinear}, with model sizes from {\bf 100M} to {\bf 1B} parameters;
    \item \textbf{Datasets:} FineWeb, C4, and OpenWebText;
    \item \textbf{Optimizers:} AdamW, Muon, and Signum.
\end{itemize}

Experimental details and complete results are provided in Appendices~\ref{appendix:exp_datails} and~\ref{appendix:collapse_exp}, respectively. Across the 26 ELR-matched comparisons summarized in Table~\ref{tab:collapse_evidence_map}, the median collapse error is \(2.5\times10^{-3}\), and all comparisons remain below \(5\times10^{-3}\). Comparable collapse precision is also observed for ViTs trained on ImageNet in Appendix~\ref{app:vit-elr-collapse}. Together, these results demonstrate the robustness of ELR collapse across architectures, model scales, datasets, and optimizers.


\vspace{-.5em}
\subsection{When Is ELR Collapse Precise?}
\label{sec:conditions_for_elr_collapse}

As discussed in the introduction and detailed in Appendix~\ref{appendix:scale-invariance}, transformers are not exactly scale invariant, so high-precision ELR collapse does not follow from an exact architectural symmetry. We therefore explored a range of model parameterizations and training configurations to identify which factors control ELR collapse. Two sensitivities emerged consistently: normalization design and the timescale of the LR--norm realization.

\vspace{-.5em}
\paragraph{QK-Norm and learnable RMSNorm gains unexpectedly improve collapse precision.}
Although QK-Norm is not required for stable training in our settings, our exploration across a broad range of module configurations revealed that it substantially improves collapse precision.
 Our default models therefore apply QK-Norm~\citep{henry2020querykey} after the query and key projections and RMSNorm~\citep{zhang2019rmsnorm}, with learnable gains, before the attention and FFN blocks.  To assess their roles, we repeat the four-schedule prescribed-ELR protocol used in Figures~\ref{fig:elr_collapse} and~\ref{fig: collapse-error-dynamics} under three nested normalization configurations: the default model, the model without QK-Norm, and the model without QK-Norm and with fixed RMSNorm gains.

\begin{figure}[!h]
    \centering
    \includegraphics[height=0.29\textwidth]{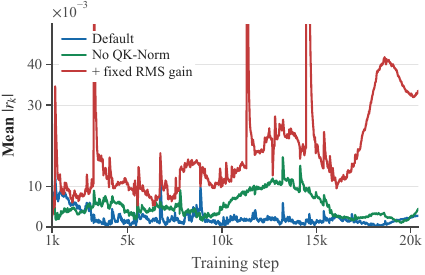}
    \hspace{1.5em}
    \includegraphics[height=0.29\textwidth]{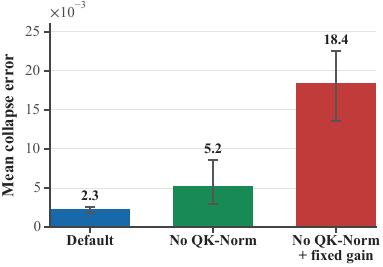}
\vspace{-.5em}
\caption{\textbf{QK-Norm and learnable RMSNorm gains sharpen ELR collapse.} Experiments use Llama-124M trained on FineWeb with AdamW. Starting from the default configuration, the ablations sequentially remove QK-Norm and then fix RMSNorm gains. \textbf{Left:} mean absolute loss residual at each step, averaged over the three ELR-matched schedules. \textbf{Right:} Mean collapse error; bars show the mean across schedules, and error bars show their range. Fixing the RMSNorm gains produces the largest error despite making the parameterization more scale invariant. In the plot labels, RMS denotes RMSNorm.
}
\label{fig:normalization_component_conditions}
\end{figure}

Figure~\ref{fig:normalization_component_conditions} shows that removing QK-Norm increases the mean collapse error from \(2.3\times10^{-3}\) to \(5.2\times10^{-3}\). Starting from the model without QK-Norm, fixing the remaining RMSNorm gains further increases the error to \(1.84\times10^{-2}\). The combined ablation therefore increases the collapse error by nearly an order of magnitude. Thus, both QK-Norm and learnable RMSNorm gains substantially improve collapse precision, although neither is designed for this purpose.

The gain ablation is particularly counterintuitive. Fixing the gains makes the parameterization more scale invariant, yet increases the collapse error by a factor of approximately \(3.5\) relative to the learnable-gain counterpart. High-precision ELR collapse therefore cannot be explained solely by static architectural scale invariance. Instead, this result suggests an additional dynamical role for normalization and gain adaptation, whose mechanism remains to be understood.


\paragraph{Faster LR--norm variation degrades ELR collapse.}
Starting from a WSD LR schedule \(\eta_k^{\mathrm{WSD}}\), we introduce the sinusoidal modulation
\[
\eta_k^{(m)}=s_m(k)\eta_k^{\mathrm{WSD}}, \qquad s_m(k)=1+0.5\sin\left(\frac{2\pi m k}{N}\right),
\]
where \(N\) is the training horizon and \(m\) controls the number of oscillation cycles. For each \(m\), we multiply the target-norm schedule by the same factor \(s_m(k)\), so that the prescribed ELR remains unchanged. Varying \(m\) therefore changes the timescale of the LR--norm realization while preserving the ELR schedule.

\begin{figure}[H]
    \centering
    \makebox[0.94\textwidth][c]{%
    \includegraphics[width=0.36\textwidth]{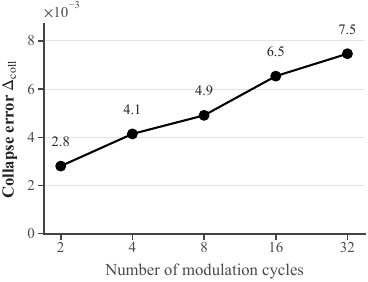}
      \hspace{0.5em}%
        \includegraphics[height=0.29\textwidth]{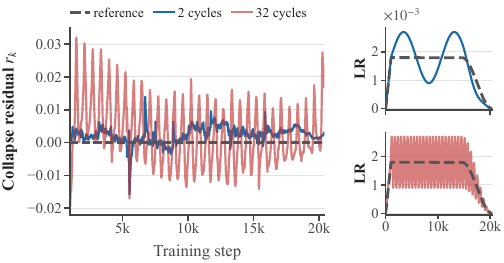}%
    }
\caption{\textbf{Rapid LR--norm variation degrades ELR collapse.} Experiments use Llama-124M trained on FineWeb with AdamW.  \textbf{Left:} The collapse error $\Delta_{\collapse}$ increases monotonically with the number of modulation cycles. \textbf{Right:} Loss residuals for representative 2- and 32-cycle modulations. The right insets show the corresponding LR schedules together with the WSD reference. In each run, the target-norm schedule is modulated by the same factor as the LR, preserving the prescribed ELR schedule. Faster modulation produces larger oscillatory residuals at the imposed frequency.
}
    \label{fig:wave_frequency_condition}
    \vspace{-.5em}
\end{figure}

Figure~\ref{fig:wave_frequency_condition}(left) shows that, over the tested range, the mean collapse error increases monotonically from \(2.8\times10^{-3}\) at two cycles to \(7.5\times10^{-3}\) at 32 cycles. Figure~\ref{fig:wave_frequency_condition}(right) reveals that the residuals oscillate at the imposed frequency, with substantially larger amplitude under faster modulation. The degradation is therefore primarily a short-timescale response rather than a systematic divergence of the loss trajectories.

Together, these ablations show that ELR is an accurate but not exact coordinate for loss dynamics: collapse precision depends on normalization design and the timescale of joint LR--norm variation. Its origin must therefore involve training dynamics beyond static scale invariance, but the underlying mechanism remains open.


\vspace{-.5em}
\section{Weight Decay and Hyperball Act through ELR}
\label{sec:weight_decay_elr}
\vspace{-.5em}

The preceding experiments establish ELR collapse using prescribed LR and norm schedules. We now test whether ELR also mediates the effects of two practical norm-control methods: weight decay and Hyperball. For each method, we train a norm-controlled target run and record the ELR trajectory it induces. We then train a comparison run under a different norm-control setting, leaving its norm trajectory unprescribed and adapting only its LR to match the target ELR. If this LR-only intervention recovers the target loss dynamics despite different norm trajectories and norm-control mechanisms, then the method's effect on loss is mediated by the ELR it induces. The optimizer-specific matching rules are given in Appendix~\ref{sub:elr_matching_protocols}.

\begin{figure}[!ht]
    \centering
    \includegraphics[width=0.8\textwidth]{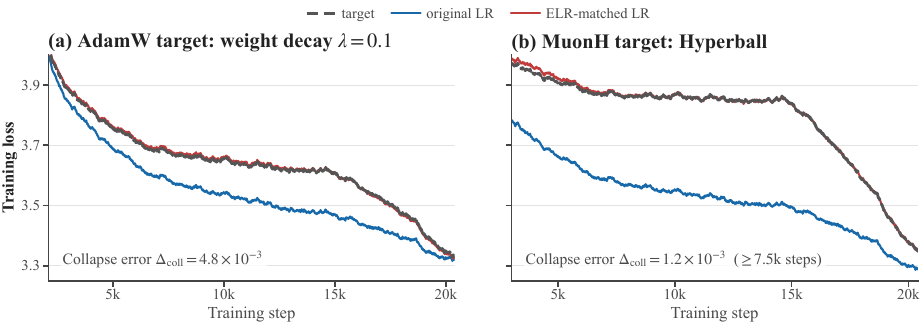}
 \vspace{-.5em}
    \caption{
    \textbf{Weight decay and Hyperball affect loss dynamics through ELR.} Experiments use Llama-124M trained on FineWeb. 
    Dashed gray curves denote the norm-controlled targets, blue curves use the original LR, and red curves adapt the LR to match the target ELR.
    \textbf{(a)} An AdamW run with weight decay \(\lambda=0.1\) serves as the target. Removing weight decay changes the loss dynamics under the original LR, whereas ELR matching recovers the target loss with a collapse error of \(4.8\times10^{-3}\).
    \textbf{(b)} A  MuonH run serves as the target. Replacing Hyperball with weight decay changes the loss dynamics under the original LR, whereas ELR matching reduces the collapse error (steps \(\geq 7.5\mathrm{k}\)) to \(1.2\times10^{-3}\).
    }
    \label{fig:wd_switch_lrnorm}
    \vspace{-1em}
\end{figure}

\vspace{-.5em}
\begin{itemize}
\item {\bf Weight decay.}
We take an AdamW run with \(\lambda=0.1\) as the target and compare it with a run using \(\lambda=0\). Under the original LR schedule, removing weight decay changes the norm evolution and causes the loss trajectory to deviate substantially from the target. When we instead adapt only the LR of the \(\lambda=0\) run to match the target ELR, its entire loss trajectory closely follows the target throughout training, with a mean collapse error of \(4.8\times10^{-3}\); see Figure~\ref{fig:wd_switch_lrnorm}(a).

\item {\bf Hyperball.}
We take MuonH (Muon with Hyperball) as the target and adapt only the LR of a MuonW run (Muon with weight decay) to match the target ELR. Figure~\ref{fig:wd_switch_lrnorm}(b) shows that MuonW with its original LR deviates substantially from MuonH, whereas ELR matching brings the loss trajectories into close alignment after a brief initial transient. This transient occurs at the start of matching, when the adapted LR is largest. We therefore evaluate collapse from step \(7.5\mathrm{k}\) onward, obtaining a mean error of only \(1.2\times10^{-3}\).
\end{itemize}

Appendix~\ref{sec:additional_results_for_section_ref_sec_weight_decay_elr} provides full trajectory diagnostics for both experiments and reports the reverse-direction interventions, with each norm-control setting serving in turn as the target. These reverse interventions yield the same conclusion.

Together, these LR-only interventions show that weight decay and Hyperball affect loss dynamics primarily through the ELR trajectories they induce. Apart from the brief Hyperball onset transient, matching ELR reduces the loss discrepancies between different norm-control mechanisms to the \(10^{-3}\) scale. This does not imply that norm control necessarily improves convergence: its effect depends on the ELR trajectory it induces. 

\section{ELR Enables Scaling-Law Transfer}
\label{sec:elr-fsl}

\vspace{-.5em}
Pairwise ELR matching shows that individual loss trajectories can be aligned across different norm-control methods. This does not yet establish that these methods obey a common predictive law. We therefore ask a stronger question: can a scaling law fitted without Hyperball transfer to Hyperball when expressed in ELR coordinates?

\vspace{-.5em}
\paragraph{Functional scaling laws.}
The functional scaling law (FSL) of \citet{li2025fsl} models loss dynamics under a general LR schedule. Given an LR schedule \(\boldsymbol{\eta}=(\eta_0,\ldots,\eta_N)\), define the intrinsic training time
$
    t_k
    \coloneqq
    \sum_{j=0}^{k}\eta_j.
$
FSL models the loss at step \(k\) as
\begin{equation}
    \label{eqn:fsl}
    L_k
    =
    L_\infty + S(t_k) + N_k,
    \qquad
    N_k
    =
    \sum_{j=0}^{k} K(t_k-t_j)\eta_j^2,
\end{equation}
where \(L_\infty\) is the irreducible loss. For large \(t\), the signal term and memory kernel satisfy \(S(t)\asymp t^{-s}\) and \(K(t)\asymp t^{-\gamma}\) for some \(s,\gamma>0\). The signal term decreases with the accumulated training time, whereas \(N_k\) captures the optimization noise accumulated along the trajectory. Increasing the LR therefore accelerates signal learning by advancing the intrinsic time, but also increases noise injection through the factor \(\eta_j^2\). We refer to this original LR-based formulation as \textbf{lr-FSL}.

Motivated by ELR collapse, we define \textbf{elr-FSL} by replacing \(\eta_j\) with \(\eta_j^{\eff}\) throughout Equation~\eqref{eqn:fsl}, including both the intrinsic training time and the noise term, while leaving the functional form unchanged. The two formulations therefore differ only in whether loss dynamics are parameterized by LR or ELR. In both cases, the realized LR or ELR schedule is supplied as input. We test scaling-law transfer across norm-control methods in this section and use elr-FSL again in Section~\ref{sec:delayed_acceleration} to analyze delayed acceleration. 


\vspace{-.5em}
\paragraph{Transfer across norm-control methods.}
We train a 124M-parameter Llama model on FineWeb under ten configurations: four without weight decay, four with weight-decay coefficient \(\lambda=0.1\), and two with Hyperball. Both FSL variants are fitted on the same four non-Hyperball trajectories. Four additional non-Hyperball trajectories evaluate generalization to held-out LR schedules under norm-control methods represented in the fitting set; we refer to these as in-distribution (ID) runs. The two Hyperball trajectories test out-of-distribution (OOD) transfer to a norm-control method absent from the fitting set. Appendix~\ref{app:fsl-details} provides the complete data split, parameterization, and fitting procedure.

Figure~\ref{fig:fsl-generalization} summarizes the results. On the two unseen Hyperball runs, elr-FSL remains accurate without refitting, with a mean RMSE of \(0.0212\), whereas lr-FSL develops a large systematic bias and reaches \(0.2508\), a \(11.83\times\) larger error. ELR also improves prediction on the held-out ID runs, reducing the mean RMSE from \(0.0239\) to \(0.0133\). Thus, the advantage of ELR is not limited to pairwise trajectory alignment: it enables a scaling law fitted under one set of norm-control methods to transfer to another:


\begin{figure}[H]
    \centering

    \textbf{(a) Prediction error across data splits}\par\vspace{0.35em}
    \small
    \setlength{\tabcolsep}{4pt}
    \renewcommand{\arraystretch}{1.1}
 \begin{tabular}{lcccc}
    \toprule
    Evaluation
    & \# Runs
    & lr-FSL RMSE $\downarrow$
    & elr-FSL RMSE $\downarrow$
    & $\mathrm{RMSE}_{\rm lr}/\mathrm{RMSE}_{\rm elr}$ $\uparrow$ \\
    \midrule
    Fit
    & 4
    & 0.0183
    & \textbf{0.0131}
    & $1.40\times$ \\
    \hdashline
    ID: held-out
    & 4
    & 0.0239
    & \textbf{0.0133}
    & $1.80\times$ \\
    \rowcolor{red!6}
    OOD: Hyperball
    & 2
    & 0.2508
    & \textbf{0.0212}
    & \textbf{$11.83\times$} \\
    \bottomrule
\end{tabular}

    \vspace{1em}

    \begin{minipage}[t]{0.41\textwidth}
        \centering
        ~~~~~~\textbf{(b) Held-out WSD schedule}\par\vspace{-.24em}
        \includegraphics[width=\linewidth]{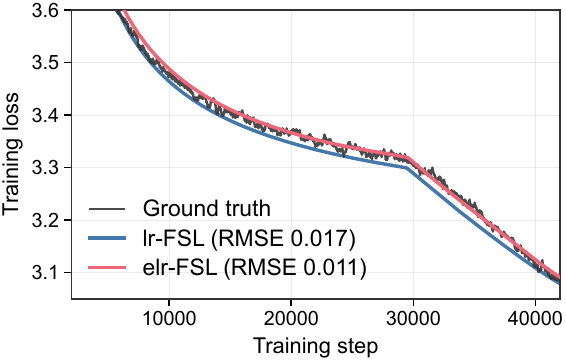}
    \end{minipage}
    \hspace{1.2em}
    \begin{minipage}[t]{0.41\textwidth}
        \centering
        ~~~~~~\textbf{(c) Transfer to Hyperball}\par\vspace{-0em}
        \includegraphics[width=\linewidth]{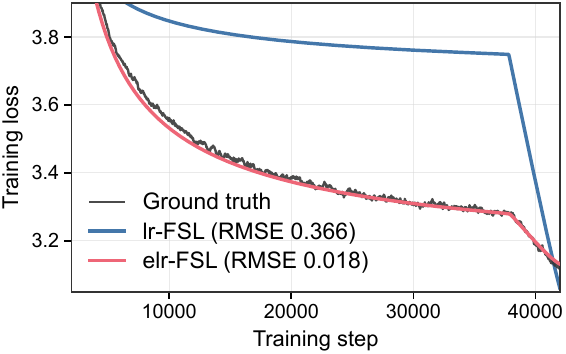}
    \end{minipage}
\caption{\textbf{ELR enables FSL to transfer to unseen Hyperball runs.} Experiments use Llama-124M trained on FineWeb with Adam under three norm-control regimes: no weight decay, decoupled weight decay, and Hyperball.  \textbf{(a)} Mean RMSE on four fitting runs, four held-out ID runs, and two OOD-Hyperball runs. 
Both FSL variants are fitted on the same four non-Hyperball curves and evaluated without refitting. The lr-FSL/elr-FSL RMSE ratio increases from \(1.8\times\) on the held-out ID runs to \(12.9\times\) on the OOD-Hyperball runs. \textbf{(b)} A representative held-out ID using weight-decay coefficient \(\lambda=0.1\). \textbf{(c)} A representative OOD-Hyperball run; no Hyperball data are used for fitting. Without refitting, elr-FSL continues to track the observed loss, whereas lr-FSL breaks down.  Table entries are averages across runs; legend values report RMSE for the individual runs shown.
}
    \label{fig:fsl-generalization}
\end{figure}

\section{Delayed Acceleration: Explanation and Control}
\label{sec:delayed_acceleration}

\begin{figure}[!h]
    \centering
    \includegraphics[width=0.92\textwidth]{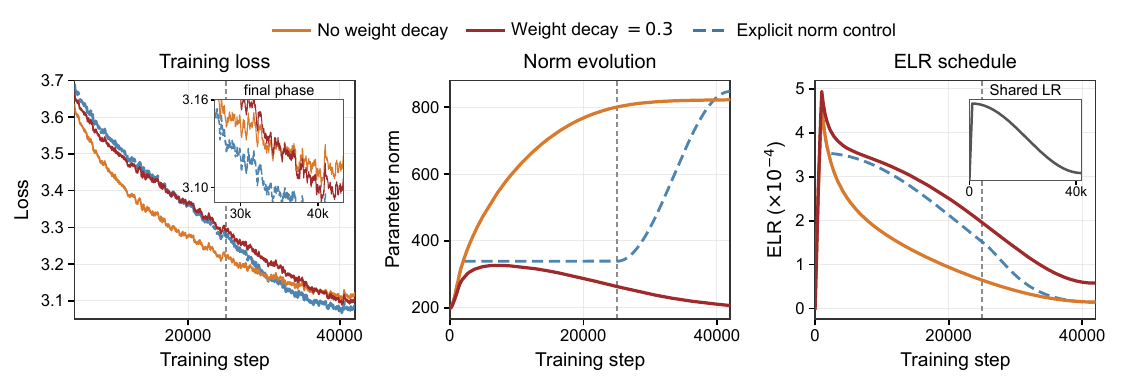}
    \caption{\textbf{ELR explains and controls delayed acceleration.} Experiments use Llama-124M trained on FineWeb with AdamW.  \textbf{Left:} Weight decay initially yields higher loss than the unregularized baseline, but overtakes it during the terminal phase. ELR-guided norm control further lowers the final loss. \textbf{Middle:} Corresponding norm evolution. Explicit norm control approximately follows the weight-decay run early, then increases the norm more rapidly after the vertical dashed line. \textbf{Right:} Corresponding ELR trajectories, with the shared LR shown in the inset. The accelerated late-stage norm growth produces a faster ELR decay.}
    \label{fig:delay-acceleration}
    \vspace{-1em}
\end{figure}

Having shown that ELR provides a common predictive coordinate across norm-control regimes, we now use it to explain a recurring late-stage benefit of norm control. Weight decay can be crucial for sustaining optimizer efficiency as LLM training scales; see \citet[Figure~2]{liu2025muon} and \citet[Figure~A7]{hoffmann2022training}. Yet its benefit often emerges only late. As shown in Figure~\ref{fig:delay-acceleration}(left), the weight-decayed run initially has higher loss than the unregularized baseline, but later overtakes it and attains a lower final loss. Hyperball exhibits the same qualitative behavior; see Appendix~\ref{appendix:delayed_acceleration}. We call this late-emerging gain \textbf{delayed acceleration} and show that it is governed by the ELR schedule induced by norm control.

\paragraph{Norm growth prematurely suppresses ELR.}
Figure~\ref{fig:delay-acceleration}(middle and right) reveals the origin of the reversal. Without weight decay, the parameter norm grows steadily, causing the ELR
$
    \eta_k^{\eff}=\eta_k/\|\bW_k\|_F
$
to decay substantially faster than the nominal LR. Weight decay restrains this norm growth and thereby sustains a larger ELR for longer. Thus, the two runs follow markedly different effective optimization trajectories despite sharing the same LR schedule. From the ELR perspective, the role of weight decay in loss dynamics is to prevent premature ELR decay.

\paragraph{The gain is acquired early and revealed late.}
The signal--noise decomposition of FSL explains why the benefit of this larger ELR is delayed. Let
\[
    t_k^{\eff}=\sum_{j=0}^{k}\eta_j^{\eff}
\]
denote the effective training time. By sustaining a larger ELR, weight decay accumulates \(t_k^{\eff}\) more rapidly and lowers the signal-learning term \(S(t_k^{\eff})\). At the same time, the larger ELR produces greater noise accumulation, which can initially mask this signal-learning advantage. As ELR decays later in training, new noise injection diminishes while previously accumulated noise continues to be forgotten. The advantage acquired earlier is then revealed as a lower training loss. Delayed acceleration therefore reflects a temporal separation: the gain is \emph{acquired early but revealed late}.

\paragraph{ELR reshaping controls delayed acceleration.}
The previous explanation suggests improving the ELR schedule induced by weight decay with \(\lambda=0.3\): preserve its relatively large ELR early in training, but reduce ELR more aggressively during the late phase. We realize this modification through a prescribed norm schedule that approximately follows the \(\lambda=0.3\) run early and grows faster later; see Figure~\ref{fig:delay-acceleration}(middle). Under the shared LR schedule, this produces a similar early ELR but a smaller late-stage ELR; see Figure~\ref{fig:delay-acceleration}(right). The resulting run preserves delayed acceleration and attains a lower final loss than the weight-decay baseline; see Figure~\ref{fig:delay-acceleration}(left). Crucially, this improvement is achieved by increasing the parameter norm more rapidly in the late phase, showing that the benefit does not arise from maintaining a uniformly smaller norm. Rather, norm control affects loss dynamics through the ELR  schedule it realizes. Although the prescribed schedule is heuristic rather than optimized, the intervention turns the ELR explanation into a control principle: delayed acceleration can be shaped directly through ELR and is not specific to weight decay itself.

Appendix~\ref{appendix:delayed_acceleration} provides additional experiments, including analogous delayed acceleration under Hyperball and its explanation through the same ELR-based mechanism.


\section{Conclusion and Discussion}
\label{sec:conclusion}
Our main result is a macroscopic empirical law for language model pretraining: LR and parameter norm govern loss dynamics primarily through their ratio, ELR. Across the regimes studied, distinct LR schedules and norm-control mechanisms produce nearly identical loss dynamics when they induce the same ELR schedule.  This reduction is analogous to the role of density in a macroscopic physical system: mass and volume may vary substantially, yet the relevant bulk behavior is governed primarily by density---their ratio---rather than by either quantity alone. Likewise, LR scheduling and norm control shape pretraining loss dynamics primarily through the ELR trajectory they jointly induce.

\paragraph{Microscopic mechanism and scope.}
ELR collapse is a macroscopic law, not a consequence of exact scale symmetry. Transformers are not exactly scale invariant, and learnable RMSNorm gains sharpen collapse even though they make the parameterization less scale invariant. Rapid LR--norm variation also produces structured oscillatory deviations at the imposed frequency. These observations suggest a dynamical compensation mechanism with a finite response timescale. A satisfactory theory should explain both why the many coupled degrees of freedom in Transformer training admit such an accurate low-dimensional description at the level of loss and why its precision depends on normalization design and the timescale of the LR--norm realization. Our claims are correspondingly limited to loss dynamics: matched loss trajectories need not imply matched parameters, representations, or downstream performance.

\paragraph{ELR as a coordinate for cross-scale transfer.}
Hyperparameter transfer is commonly posed as separate rules for the LR, weight-decay coefficient, and norm target~\citep{yang2021tuning,ghosh2026understanding,kosson2026weight}. ELR suggests a more structured two-stage problem: first determine the ELR schedule appropriate for a target scale, then choose a stable LR--norm realization. This separates two questions that current tuning practice often conflates: what effective trajectory should training follow, and how should that trajectory be realized? A useful transfer theory should determine how the desired ELR schedule changes with model size, batch size, data budget, and training horizon, while allowing the LR and norm-control mechanism to be chosen according to the constraints of each setting. Such a formulation would replace method-specific transfer heuristics with a common dynamical target.

\paragraph{ELR-first pretraining.}
Current practice often treats the LR schedule, weight decay, and norm constraints as separate design choices. Our results suggest a different abstraction: the ELR schedule is the design object~\citep{li2026optimal,wang2026fast}, whereas LR scheduling and norm control are mechanisms for realizing it. Norm control is useful not because a particular norm is intrinsically preferable, but because it expands and shapes the set of ELR schedules that training can realize. The explicit norm-control experiment makes this distinction concrete: increasing the parameter norm more rapidly late in training improves the final loss by producing a more favorable ELR decay. This does not mean that ELR captures every aspect of training; stability, numerical precision, and transferability remain additional constraints on how an ELR schedule should be realized. For loss dynamics, however, the practical principle is simple: \emph{design the ELR schedule first, then choose the LR and norm-control mechanism that realizes it robustly.}

\section*{Acknowledgement}
Lei Wu is supported by the National Natural Science Foundation of China (NSFC 12522120, NSFC 92470122, and NSFC 12288101). Zihan Liu is supported by Ant Group Research Intern Program.
We thank Shengtao Guo and Binghui Li for helpful discussions.

\bibliography{collapse}
\bibliographystyle{ims}

\newpage

\appendix
\counterwithin{figure}{section}
\counterwithin{table}{section}
\counterwithin{equation}{section}

\part{Appendix}
\parttoc

\newpage

\section{Nontriviality and Precision of ELR Collapse}
\subsection{Transformers Lack Scale Invariance}
\label{appendix:scale-invariance}

Our ELR is defined using the norm of the matrix-valued parameters. Transformers, however, also contain trainable vector-valued parameters, most notably the biases of FFN layers and gains in normalization layers. Let $\cL(\bW,\bphi)$ denote the loss, where $\bW$ collects the matrix-valued parameters and $\bphi$ the vector-valued parameters. The strongest relevant symmetry one might posit is matrix-only scale invariance:
\begin{equation}
    \cL(c\bW,\bphi)
    =
    \cL(\bW,\bphi),
    \qquad
    \forall\,c>0.
    \label{eq:matrix-scale-invariance}
\end{equation}
However, even if this symmetry held exactly, matching the  ELR would not by itself imply matching loss trajectories  since the loss is sensitive to the scale of $\bphi$.

More seriously,   the matrix-only symmetry in Eq.~\eqref{eq:matrix-scale-invariance} also fails for several architectural reasons:

\begin{itemize}
\item \textbf{Pre-norm residual blocks.}
A pre-norm residual block updates the hidden state as
\begin{equation}
    \bh^{\ell+1}
    =
    \bh^\ell
    +
    F_\ell\!\left(
        \operatorname{RMSNorm}_{\bgamma_\ell}(\bh^\ell);
        \bW_\ell
    \right).
    \label{eq:prenorm-matrix-scaling}
\end{equation}
The identity path preserves the scale of $\bh^\ell$, whereas RMSNorm removes it before the residual branch. Thus, rescaling the upstream matrices that determine $\bh^\ell$ affects the two paths differently. Rescaling $\bW_\ell$ likewise changes the residual branch relative to the identity path. Hence, the block is not invariant under a  rescaling of its matrix parameters.

    \item \textbf{Value and output projections.}
    The attention value--output path contains the product
    \[
        (\bH\bW_V)\bW_O.
    \]
    Jointly rescaling $\bW_V$ and $\bW_O$ by $c$ scales this path by $c^2$, even if the attention weights remain fixed. Its magnitude relative to the residual stream therefore changes.

    \item \textbf{SwiGLU feed-forward networks.}
    The SwiGLU FFN combines several matrix scales:
    \begin{equation}
        \operatorname{MLP}(\bH)
        =
        \left[
            \operatorname{SiLU}(\bH\bW_G)
            \odot
            (\bH\bW_U)
        \right]\bW_D.
        \label{eq:swiglu-matrix-scaling}
    \end{equation}
    Its multiplicative gate couples the scales of multiple projections and moreover, SiLU is not homogeneous. Consequently, a common rescaling of $\bW_G$, $\bW_U$, and $\bW_D$ cannot be canceled by a single global scale factor.

    \item \textbf{Token embeddings and output logits.}
Rescaling the token embeddings changes the scale of the initial residual stream, which is not eliminated by subsequent residual blocks. Rescaling the  head layer changes the logit scale and hence the effective softmax temperature, to which cross-entropy is not invariant. When the embedding and output matrices are tied, the same parameter introduces scale dependence at both ends of the network.
\end{itemize}

\subsection{Interpreting Collapse Errors at the $10^{-3}$ Scale}
\label{appendix:stochastic_baseline}

Fix a source of stochasticity \(c\), and let \(L_{i,k}\) denote the smoothed loss of run \(i\) at step \(k\). To match the definition of collapse error, we measure run-to-run variation using the pairwise mean absolute discrepancy
\[
    d_{ij}
    \coloneqq
    \frac{1}{|\mathcal T|}
    \sum_{k\in\mathcal T}
    \left|L_{i,k}-L_{j,k}\right|,
    \qquad \text{ for }1\leq i<j\leq n,
\]
where \(\mathcal T\) is the set of evaluation steps. We then define the stochastic variation associated with \(c\) by averaging over all unordered pairs:
\[
    \Delta_{\mathrm{stoch}}(c)
    \coloneqq
    \frac{2}{n(n-1)}
    \sum_{1\le i<j\le n} d_{ij}.
\]
This definition uses the same mean absolute trajectory discrepancy as \(\Delta_{\mathrm{coll}}\), allowing a direct comparison between ELR-collapse errors and ordinary run-to-run variation.

\begin{figure}[!h]
    \centering
    \includegraphics[width=1.0\textwidth]{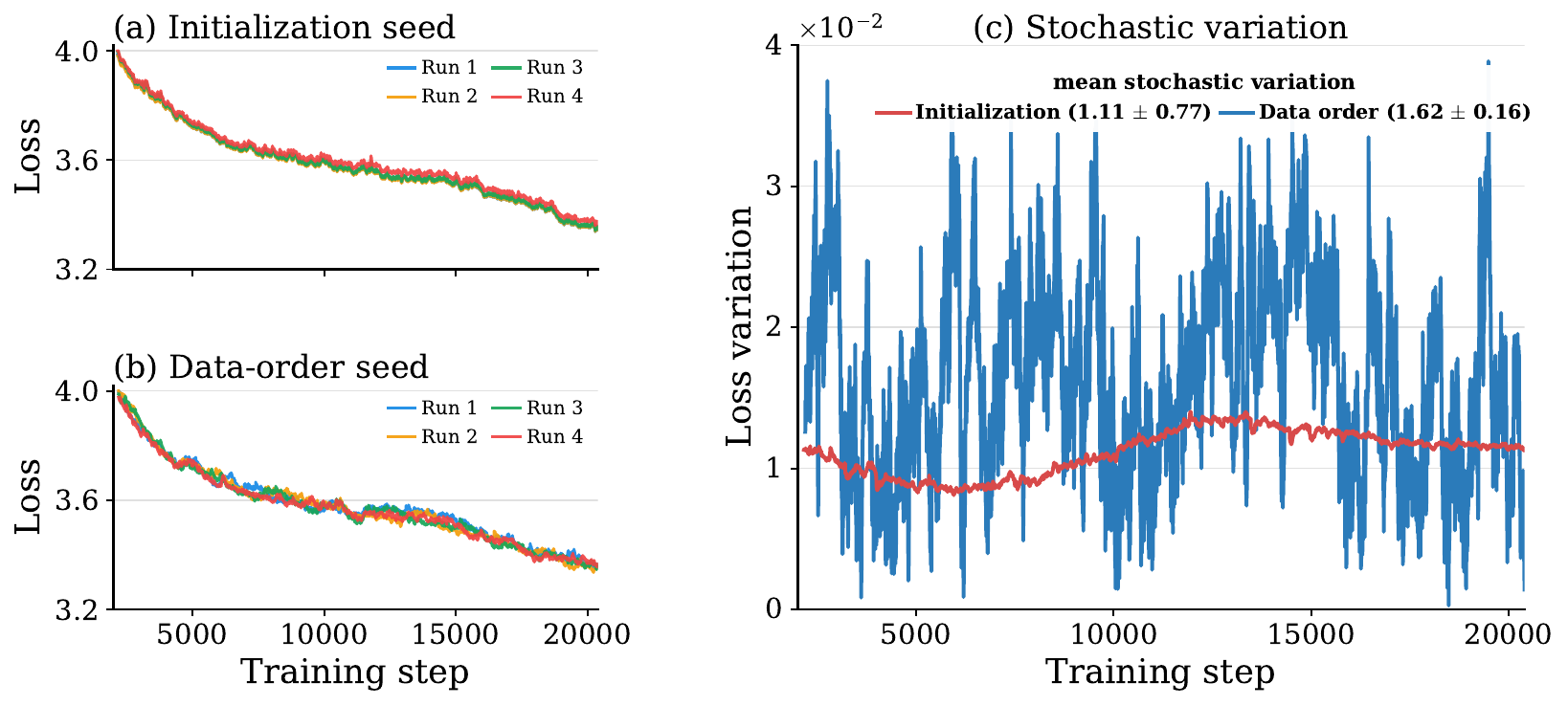}
\caption{\textbf{Run-to-run loss variation substantially exceeds ELR-collapse error.} \textbf{(a,b)} 
Training-loss trajectories from repeated runs differing only in initialization seed (a) or data-order seed (b). \textbf{(c)} The step-wise mean pairwise loss discrepancy over training. Legend values report the mean and standard deviation of the pairwise time-averaged discrepancies across unordered run pairs, in units of \(10^{-2}\). The resulting time-averaged variations are \(\Delta_{\mathrm{stoch}}(\mathrm{init})=1.11\times10^{-2}\) and \(\Delta_{\mathrm{stoch}}(\mathrm{data\text{-}order})=1.62\times10^{-2}\). By comparison, the three ELR-matched Llama-124M runs in Figures~\ref{fig:elr_collapse} and \ref{fig: collapse-error-dynamics}, with both seeds fixed, have collapse errors of \(1.8\times10^{-3}\), \(2.5\times10^{-3}\), and \(2.6\times10^{-3}\).
}
\label{fig:stochastic_variability_baseline}
\end{figure}

Figure~\ref{fig:stochastic_variability_baseline} reports run-to-run variation for Llama-124M trained on FineWeb with AdamW. Repeating the same configuration while changing only the initialization seed or data-order seed gives
\[
\boxed{
\Delta_{\mathrm{stoch}}(\text{init})=1.11\times10^{-2},
\qquad
\Delta_{\mathrm{stoch}}(\text{data-order})=1.62\times10^{-2}.
}
\]

For comparison, the three ELR-matched runs in Figure~\ref{fig:elr_collapse} under the same training configurations, which keep both seeds fixed, have collapse errors between \(1.8\times10^{-3}\) and \(2.6\times10^{-3}\). Thus, even the largest collapse error is more than four times smaller than the lower stochastic-variation baseline; across the reported values, the difference is a factor of \(4.3\)–\(9.0\).

\section{Experimental Details}
\label{appendix:exp_datails}

This appendix specifies the shared training configurations and the ELR matching protocols  used throughout the paper. Appendix~\ref{appendix:collapse_exp} then reports the corresponding trajectory-level results.

\subsection{Training Configurations}

Tables~\ref{tab:dense_model_configurations} and \ref{tab:moe_model_configurations} summarize the training configurations used. Below we provide further details.

\paragraph{Models.} We consider several Transformer architectures in our pretraining experiments.
\begin{itemize}[leftmargin=2em]
    \item {\bf Llama.} Llama~\citep{touvron2023llama} is a popular dense
    decoder-only Transformer architecture, incorporating Rotary Positional Encoding (RoPE), Swish-Gated Linear Units
    (SwiGLU), and root mean square layer normalization (RMSNorm). Our 124M and 1B parameter
    implementations are adapted from Modded-NanoGPT%
    \footnote{\url{https://github.com/KellerJordan/modded-nanogpt}}. They use tied
    embeddings, QK-Norm, and a GPT-2 tokenizer with a padded vocabulary
    of 50,304 tokens.

    \item {\bf Qwen3.} Qwen3~\citep{yang2025qwen3} is a family of
    decoder-only language models that includes both dense and
    mixture-of-experts (MoE) variants.  Our Qwen3 Dense and Qwen3-MoE
    implementations are adapted from the official Qwen3 release.  Compared
    with Llama, Qwen3 uses Grouped-Query Attention and QK-Norm.
    Our Qwen3-MoE model replaces the dense MLP with a sparse 32-expert MoE
    layer using top-4 routing.

    \item {\bf Kimi Delta Attention.} Kimi Delta Attention (KDA) is a
    gated linear-attention mechanism introduced in Kimi Linear~\citep{kimi2025kimilinear}.
    Our local implementation replaces three out of every four standard
    self-attention layers with KDA, giving a 3:1 KDA to standard attention ratio.
\end{itemize}

\paragraph{Datasets.} Models are pretrained on the following datasets:
\begin{itemize}[leftmargin=2em]
    \item {\bf FineWeb.} FineWeb~\citep{penedo2024fineweb} is a large-scale English web dataset derived from Common Crawl.
    
    \item {\bf Colossal Clean Crawled Corpus (C4)}~\citep{raffel2020exploring}.
    It is a large-scale public language dataset widely used for LLM pretraining, including T5~\citep{raffel2020exploring}.

    \item {\bf OpenWebText}~\citep{Gokaslan2019OpenWeb}.
    It is an open-source recreation of the WebText corpus and is extensively used for LLM pretraining, including RoBERTa~\citep{liu2019roberta} and nanoGPT\footnote{\url{https://github.com/karpathy/nanoGPT}}.

\end{itemize}

\begin{table}[!h]
    \centering
    \caption{\textbf{Dense model configurations and peak learning rates used in
    the ELR-collapse experiments.}  Q/KV lists the numbers of query and
    key/value heads.}
    \label{tab:dense_model_configurations}
    \small
    \setlength{\tabcolsep}{4.5pt}
    \renewcommand{\arraystretch}{1.12}
    \begin{tabular}{lccccccc}
        \toprule
        Model & size & $d_{\mathrm{model}}$ & $d_{\mathrm{FF}}$
        & Q/KV & depth & peak lr & final lr \\
        \midrule
        Llama-124M & 124M & 768 & 3072 & 6 & 12
            & $1.8\times10^{-3}$ & 0 \\
        Llama-135M & 135M & 576  & 1536 & 9/3 & 30
            & $6\times10^{-4}$ & $6\times10^{-5}$ \\
        Llama-1B & 1.07B & 2560 & 10240 & 20 & 12
            & $3\times10^{-4}$ & 0 \\
        Qwen3-145M & 145M & 768 & 2304 & 12/6 & 12
            & $1.8\times10^{-3}$ & 0 \\
        Kimi Delta Attention-127M & 127M & 768 & 3072 & 6 & 12 & $1.8\times10^{-3}$ & 0 \\
        \bottomrule
    \end{tabular}
\end{table}

\begin{table}[!tbh]
    \centering
    \caption{\textbf{Qwen3-MoE configuration and peak learning rate used in
    the ELR-collapse experiment.} Q/KV lists the numbers of query and
    key/value heads; top-$k$ gives the number of active experts per token.}
    \label{tab:moe_model_configurations}
    \small
    \setlength{\tabcolsep}{4.5pt}
    \renewcommand{\arraystretch}{1.12}
    \begin{tabular}{lccccccccc}
        \toprule
        Model & size & $d_{\mathrm{model}}$ & $d_{\mathrm{FF}}$
        & Q/KV & depth & $n_{\mathrm{experts}}$ & top-$k$ & peak lr
        & final lr \\
        \midrule
        Qwen3-MoE & 586M & 768 & 576 & 12/4 & 12 & 32 & 4
            & $1.8\times10^{-3}$ & 0 \\
        \bottomrule
    \end{tabular}
\end{table}

\paragraph{Training recipe.}
Unless stated otherwise, we use the following configuration throughout. All runs within each comparison share the same initialization and data-order seeds, eliminating the seed-induced variability discussed in Appendix~\ref{appendix:stochastic_baseline}. We use AdamW with $\beta_1=0.9$ and $\beta_2=0.95$, the MoonShot variant of Muon~\citep{liu2025muon} with momentum coefficient $\beta=0.95$, and Signum~\citep{bernstein2018signsgd} with momentum coefficient $\beta=0.95$.  Unless otherwise specified, weight decay is set to zero; nonzero weight decay is used only when explicitly indicated. The default LR schedule is warmup--stable--decay (WSD)~\citep{hu2024minicpm}. Unless otherwise specified, all language model experiments use batch size $B=128$ and sequence length $1024$.

\paragraph{Loss smoothing.}
For all collapse and ELR-matching experiments, we apply an exponential moving average with \(\alpha_{\mathrm{EMA}}=0.99\) to the raw loss at each training step. The smoothing is performed independently for each run before computing the loss residual \(r_k\) and mean collapse error \(\Delta_{\mathrm{coll}}\). Both the plotted loss curves and the reported collapse errors are therefore based on the same smoothed losses. Unless otherwise stated, \(\mathcal T\) comprises all post-warmup training steps.

\vspace{-0.5em}
\subsection{ELR-Matching Protocols}
\label{sub:elr_matching_protocols}

The conversion from LR to ELR depends on how the optimizer parameterizes its update. We write \(\bW_k\) for a matrix-valued parameter immediately before training step \(k\), \(\bU_k\) for the unscaled optimizer update, and \(\eta_k\) for the LR. When \(\eta_k\) directly multiplies \(\bU_k\), the ELR is
\[
\eta_k^{\eff}=\frac{\eta_k}{\|\bW_k\|_F}.
\]

However, Hyperball requires a different conversion because it normalizes the update:
\[
\bW_{k+1}
=
\texttt{Norm}_R\!\left(
\bW_k-\eta_k\texttt{Norm}_R(\bU_k)
\right),
\qquad
\texttt{Norm}_R(\bQ)
\coloneqq
R\frac{\bQ}{\|\bQ\|_F}.
\]
The inner update can be written as
\[
\bW_k-\eta_k\texttt{Norm}_R(\bU_k)
=
\bW_k-\frac{\eta_kR}{\|\bU_k\|_F}\bU_k.
\]
Because Hyperball maintains \(\|\bW_k\|_F=R\), its ELR is therefore
\[
\eta_k^{\eff}=\frac{\eta_k R/\|\bU_k\|_F}{R}=\frac{\eta_k}{\|\bU_k\|_F}.
\]
We use these optimizer-specific definitions in both matching protocols below.

\paragraph{Intervention timing.}
Let \(k_0\) denote the first training step after warmup. All matching controls and interventions are activated at \(k_0\). Before \(k_0\), the compared runs follow the same warmup configuration and enter the controlled phase from the same parameter state. All ELR schedules below refer to the post-warmup phase \(k\geq k_0\).

We use two matching protocols. The first explicitly controls the LR and parameter norm  schedule to realize a prescribed ELR schedule. The second adapts only the LR to reproduce the ELR induced by a reference run.

\paragraph{Matching a prescribed ELR schedule.}
Given a target ELR schedule \(\{\eta_k^{\eff}\}_{k\geq k_0}\), we construct multiple runs with different LR and norm schedules. For an optimizer whose LR directly multiplies its update, we choose an LR schedule \(\{\eta_k\}_{k\geq k_0}\) and define the corresponding norm schedule by
\[
\rho_k\coloneqq\frac{\eta_k}{\eta_k^{\eff}}.
\]
We scale the LR schedule so that \(\rho_{k_0}=\|\bW_{k_0}\|_F\), ensuring continuity at the start of the controlled phase. At step \(k\), we first apply the underlying optimizer update with LR \(\eta_k\), obtaining a provisional parameter \(\widetilde{\bW}_{k+1}\), and then project it onto the next target norm:
\[
\bW_{k+1}
\gets
\rho_{k+1}
\frac{\widetilde{\bW}_{k+1}}
{\|\widetilde{\bW}_{k+1}\|_F}.
\]
Consequently, \(\|\bW_k\|_F=\rho_k\) and
$
\frac{\eta_k}{\|\bW_k\|_F}
=
\frac{\eta_k}{\rho_k}
=
\eta_k^{\eff}
$
throughout the controlled phase. Repeating this construction with different LR schedules produces distinct LR and norm trajectories that realize the same prescribed ELR schedule.

For a Hyperball run, its prescribed radius supplies the norm control. After computing \(\bU_k\), we instead set
\[
\eta_k=\eta_k^{\eff}\|\bU_k\|_F,
\]
which realizes the same prescribed ELR under the Hyperball definition above.

\paragraph{Reference-run ELR matching.}
This protocol does not prescribe the norm trajectory of the matched run. Its optimizer and norm-control mechanism are left unchanged, and only its LR is adapted. This allows us to test whether a norm-control method affects loss dynamics through the ELR it induces.

Let run A be the reference run and run B the matched run. We first train run A and record its ELR using the appropriate definition:
\[
\eta_k^{\eff,A}
=
\begin{cases}
\eta_k^A/\|\bW_k^A\|_F, & \text{if run A uses a direct-update optimizer},\\[2pt]
\eta_k^A/\|\bU_k^A\|_F, & \text{if run A uses Hyperball}.
\end{cases}
\]
When training run B, we set
\[
\eta_k^B
=
\begin{cases}
\eta_k^{\eff,A}\|\bW_k^B\|_F, & \text{if run B uses a direct-update optimizer},\\[2pt]
\eta_k^{\eff,A}\|\bU_k^B\|_F, & \text{if run B uses Hyperball}.
\end{cases}
\]
Thus, the matched run tracks the reference ELR while retaining its own norm evolution and norm-control mechanism.

\section{Quantitative Validation (Supplement to Section~\ref{sec:elr_state_variable})}
\label{appendix:collapse_exp}

Appendix~\ref{appendix:exp_datails} specifies the shared training configurations and ELR-matching protocols used throughout the experiments. For a matched run with smoothed loss $L_k$ and a reference run with smoothed loss $L_k^{\mathrm{ref}}$, we quantify collapse precision using the collapse residual and mean collapse error introduced in Section~\ref{sec:preliminaries}:
\[
r_k:=L_k-L_k^{\mathrm{ref}},
\qquad
\Delta_{\mathrm{coll}}
:=\frac{1}{|\cT|}
\sum_{k\in\cT}|r_k|,
\]
where \(\mathcal T\) denotes the set of training steps over which collapse is evaluated. We use these quantities throughout this appendix to report both the trajectory-level deviations and their average magnitude.

Table~\ref{tab:collapse_evidence_map} summarizes the mean collapse errors and principal configurations for all prescribed-ELR validation experiments, including those presented in Section~\ref{sec:quantitative_validation}. Unless otherwise noted, each setting compares three LR schedules---linear-up, linear-down, and sinusoidal---with the target-norm schedule adjusted so that all runs realize the same prescribed ELR schedule.

\newcolumntype{Y}{>{\raggedright\arraybackslash}X}

\begin{table}[!h]
    \centering
    \caption{\textbf{Collapse errors across prescribed-ELR validation settings.} Each row compares runs with different LR and parameter-norm schedules that realize the same prescribed ELR schedule. Entries report the mean collapse error $\Delta_{\collapse}$, in units of $10^{-3}$, for linear-up, linear-down, and sinusoidal LR schedules. An em dash denotes a schedule that was not evaluated. 
    }
    \label{tab:collapse_evidence_map}

    \small
    \setlength{\tabcolsep}{2.5pt}
    \renewcommand{\arraystretch}{1.10}

    \begin{tabular}{@{}c l l l *{3}{c} l@{}}
        \toprule
        \multirow{2}{*}{No.}
        & \multirow{2}{*}{Validation setting}
        & \multirow{2}{*}{Model}
        & \multirow{2}{*}{Dataset}
        & \multicolumn{3}{c}{$\Delta_{\collapse}\;(\times 10^{-3})$}
        & \multirow{2}{*}{Figure} \\
        \cmidrule(lr){5-7}
        & & & & Up & Down & Sine & \\
        \midrule

        \multicolumn{8}{@{}l}{\textit{Experiments reported in the main text}} \\
        \addlinespace[1pt]

        1 & Dense Llama & Llama-124M & FineWeb & $1.8$ & $2.5$ & $2.6$ & Fig.~\ref{fig:elr_collapse} \\
        2 & MoE & Qwen3-MoE 586M & FineWeb & $3.0$ & $4.1$ & $3.3$ & Fig.~\ref{fig:elr_collapse} \\

        \midrule
        \multicolumn{8}{@{}l}{\textit{Additional experiments reported in the appendix}} \\
        \addlinespace[1pt]

        3 & 1B scale & Llama-1B & FineWeb & $2.0$ & $2.0$ & \textemdash & Fig.~\ref{fig:app:llama1b_full_trajectory} \\
        4 & Qwen3 Dense & Qwen3-145M & FineWeb & $1.2$ & $1.9$ & $2.7$ & Fig.~\ref{fig:app:validation_diagnostics}(a) \\
        5 & Linear attention & KDA-127M & FineWeb & $0.9$ & $1.7$ & $3.2$ & Fig.~\ref{fig:app:validation_diagnostics}(b) \\
        6 & C4 & Qwen3-145M & C4 & $2.1$ & $2.1$ & $2.5$ & Fig.~\ref{fig:app:validation_diagnostics}(c) \\
        7 & OpenWebText & Qwen3-145M & OpenWebText & $3.5$ & $3.4$ & $3.5$ & Fig.~\ref{fig:app:validation_diagnostics}(d) \\
        8 & Signum & Llama-124M & FineWeb & $4.8$ & $1.6$ & $4.9$ & Fig.~\ref{fig:app:validation_diagnostics}(e) \\
        9 & Muon & Llama-124M & FineWeb & $1.3$ & $1.8$ & $2.5$ & Fig.~\ref{fig:app:validation_diagnostics}(f) \\

        \bottomrule
    \end{tabular}
\end{table}

\begin{figure}[H]
    \centering
    \includegraphics[width=0.7\textwidth]{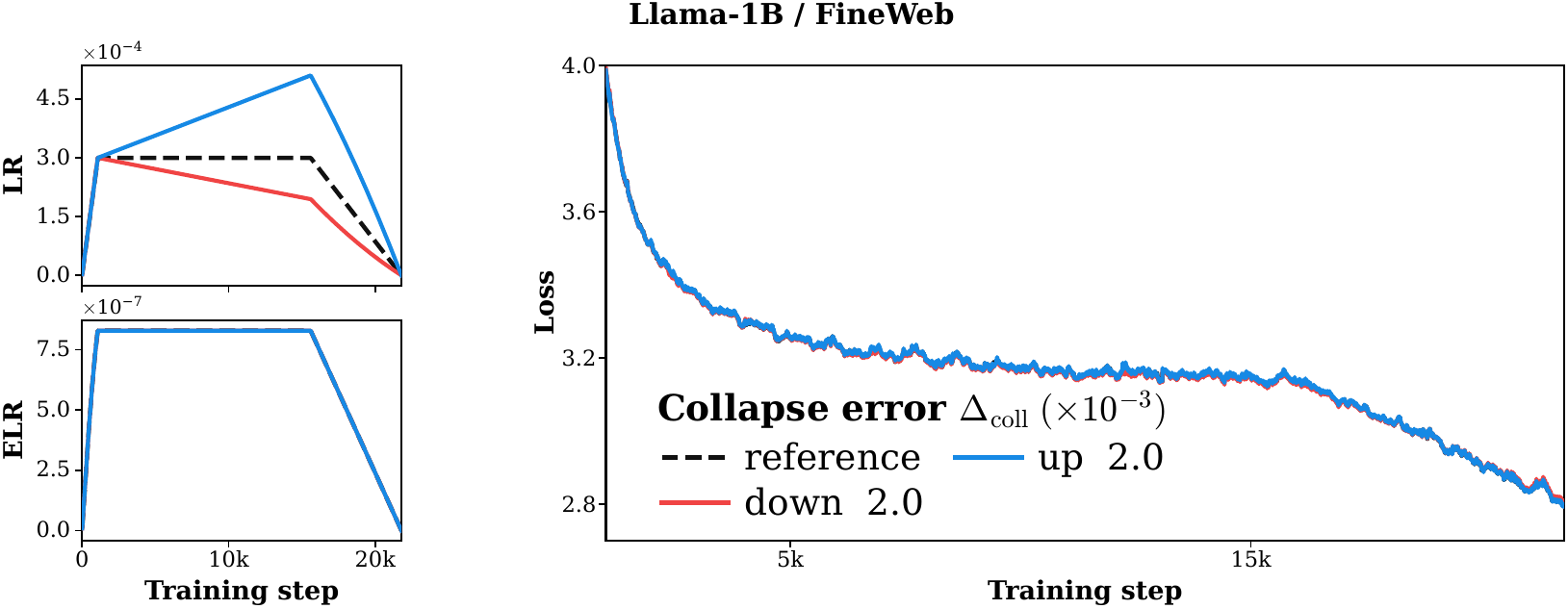}
    \includegraphics[width=0.7\textwidth]{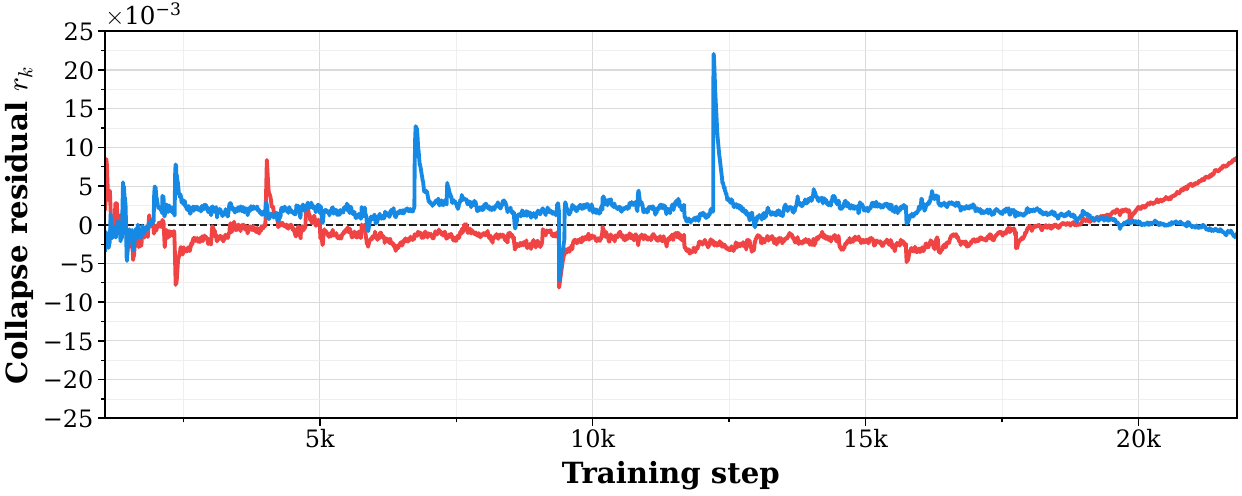}
    \caption{\textbf{ELR collapse in Llama-1B on FineWeb.}
    The compact panels show LR and ELR. The loss panel
    compares the two varied schedules with the constant reference and reports their
    mean absolute collapse errors.}
    \vspace{-1em}
    \label{fig:app:llama1b_full_trajectory}
\end{figure}

\paragraph{1B-scale stress test.}
As a scale stress test, we repeat the prescribed-ELR experiment with a Llama-1B model trained on FineWeb. Figure~\ref{fig:app:llama1b_full_trajectory} reports the full LR, ELR, loss, and collapse residual trajectories for the linear-up and linear-down realizations. Each yields a mean collapse error of \(2.0\times10^{-3}\), showing that high-precision ELR collapse persists at the 1B scale.

\begin{figure}[!h]
    \centering
    \includegraphics[width=1.0\textwidth]{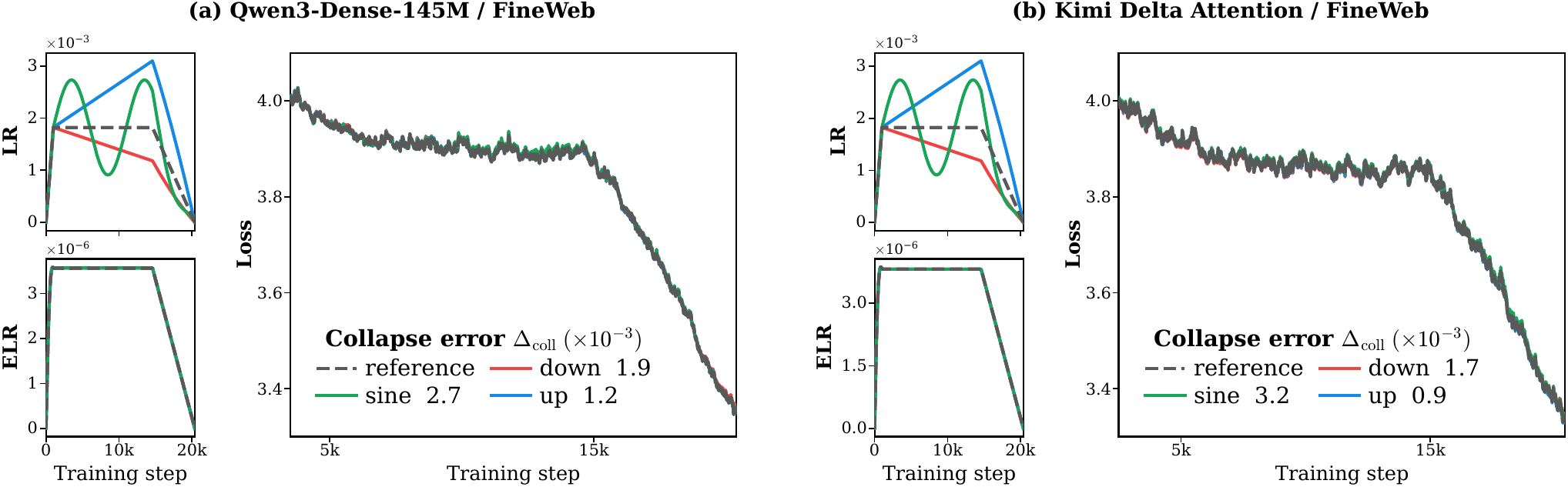}
    \\[0.5em]
    \includegraphics[width=1.0\textwidth]{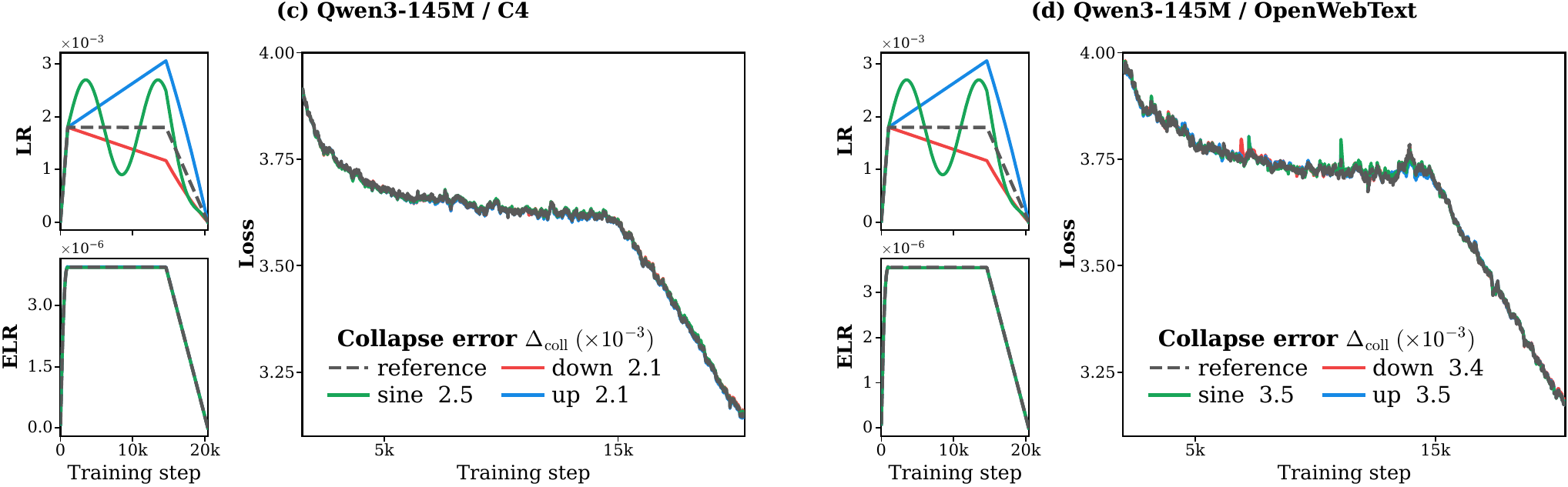}
    \\[0.5em]
    \includegraphics[width=1.0\textwidth]{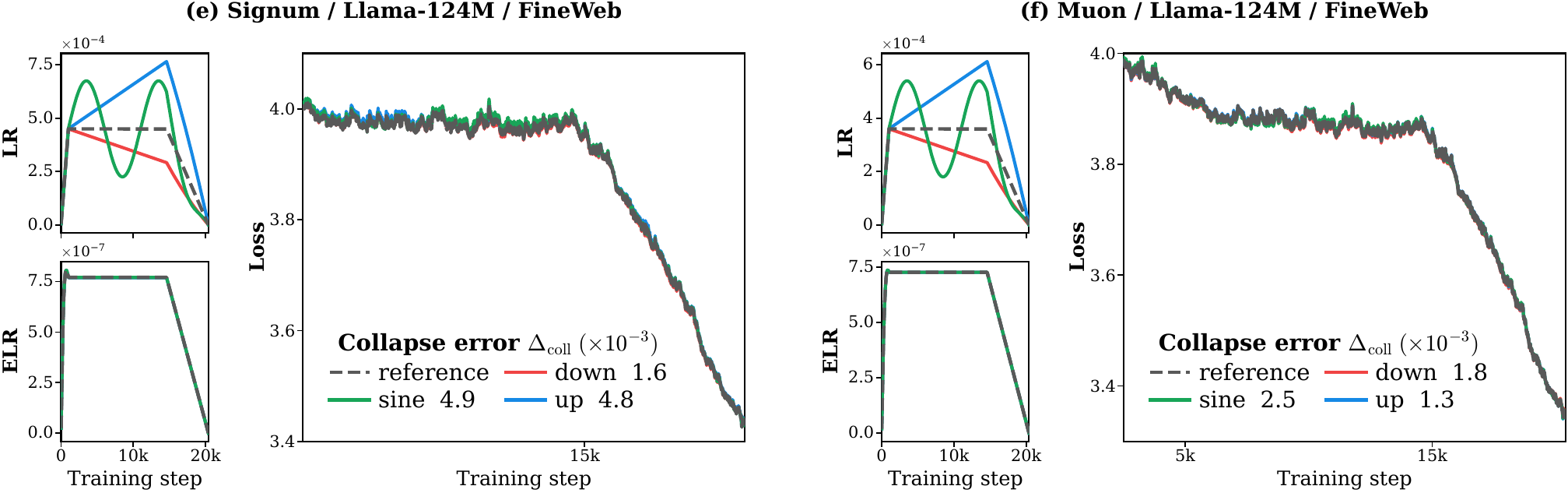}
\caption{\textbf{Additional validation of ELR collapse across training configurations.} \textbf{(a)} Replicates the prescribed-ELR experiment with Qwen3-145M on FineWeb. \textbf{(b)} Applies the same protocol to Kimi Delta Attention (KDA). \textbf{(c,d)} Repeat the Qwen3-145M experiment on C4 and OpenWebText, respectively. \textbf{(e,f)} Evaluate Signum and Muon with Llama-124M, respectively. Together, these experiments test ELR collapse across model architectures, pretraining datasets, and optimizers. In each subfigure, the loss-panel legend reports the mean absolute collapse error relative to the constant-schedule reference run.}
    \label{fig:app:validation_diagnostics}
\end{figure}

\paragraph{Broader validation.}
Figure~\ref{fig:app:validation_diagnostics} reports trajectory-level diagnostics for the remaining experiments. These extend the validation to dense Qwen3 and Kimi Delta Attention, to C4 and OpenWebText, and to the Signum and Muon optimizers. Each panel follows the LR--ELR--loss layout of Figure~\ref{fig:elr_collapse}.

\FloatBarrier
\section{Weight Decay and Hyperball (Supplement to Section~\ref{sec:weight_decay_elr})}
\label{sec:additional_results_for_section_ref_sec_weight_decay_elr}

This section reports the LR and ELR trajectories underlying the intervention experiments in Section~\ref{sec:weight_decay_elr} and repeats each intervention with the matching direction reversed. All adapted LRs follow the reference-run ELR-matching protocol in Appendix~\ref{sub:elr_matching_protocols}.

\begin{figure}[!htbp]
    \centering
    \includegraphics[width=1.0\textwidth]{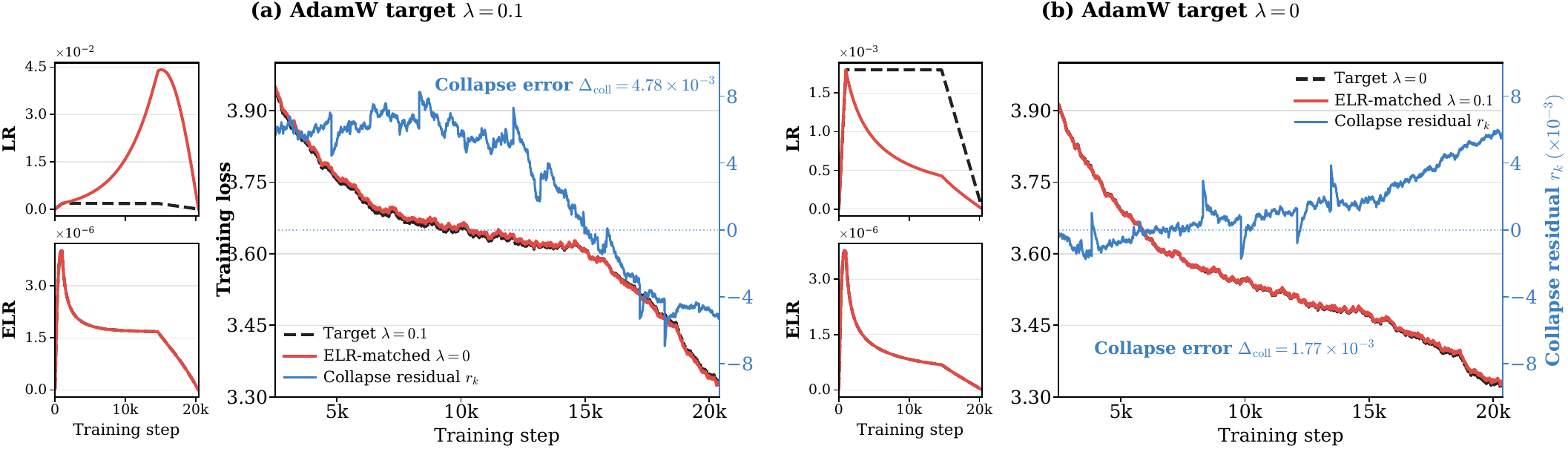}
\caption{\textbf{LR-only ELR matching recovers loss dynamics across AdamW weight-decay settings.} Experiments use Llama-124M trained on FineWeb with AdamW. In each subfigure, only the LR of the comparison run is adapted to match the target ELR, while its parameter norm evolves freely. The compact panels show the LR and ELR trajectories; the large panel compares the losses and reports the mean collapse error. \textbf{(a)} The run with weight decay \(\lambda=0.1\) serves as the target, and the LR of the \(\lambda=0\) run is adapted. \textbf{(b)} The matching direction is reversed: the \(\lambda=0\) run serves as the target, and the LR of the \(\lambda=0.1\) run is adapted. The loss trajectories remain closely aligned in both directions.}
\label{fig:app:act_adamw_diagnostics}
\end{figure}

\paragraph{Weight decay with AdamW.}
Figure~\ref{fig:app:act_adamw_diagnostics}(a) supplements Figure~\ref{fig:wd_switch_lrnorm}(a) with the corresponding LR and ELR trajectories. The run with \(\lambda=0.1\) serves as the target, while the LR of the run without weight decay is adapted to track its ELR schedule. Because the norm of the \(\lambda=0\) run grows more rapidly, its adapted LR increases through most of training and reaches approximately \(4\times10^{-2}\) before the terminal decay. Figure~\ref{fig:app:act_adamw_diagnostics}(b) reverses the direction: the \(\lambda=0\) run serves as the target, and the LR of the \(\lambda=0.1\) run is adapted accordingly. The loss trajectories remain closely aligned in both directions.

\begin{figure}[!h]
    \centering
    \includegraphics[width=1.0\textwidth]{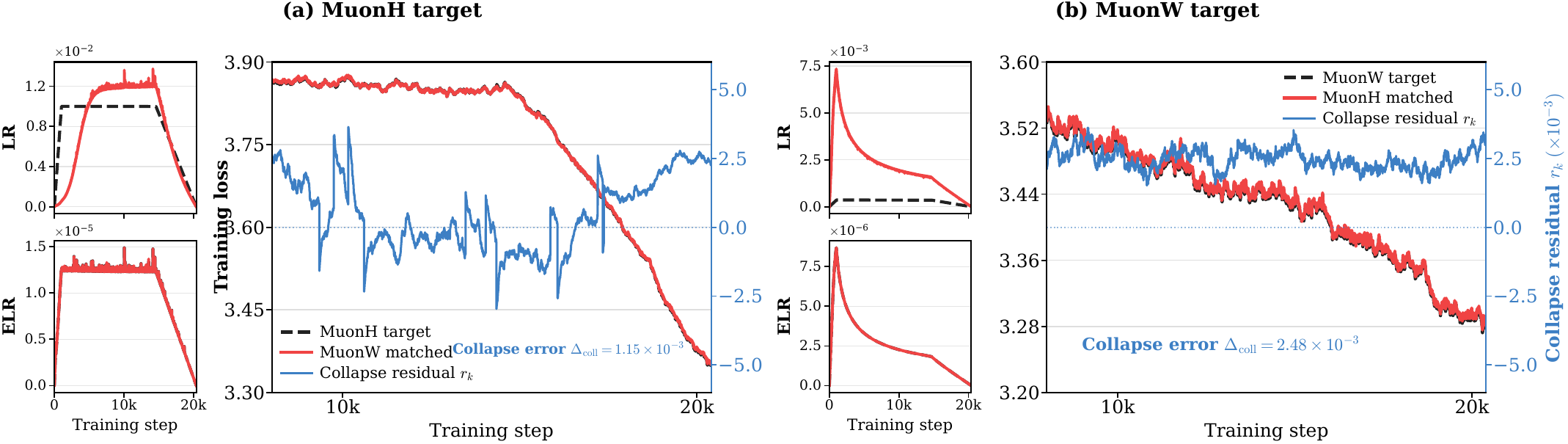}
\caption{\textbf{LR-only ELR matching aligns MuonW and MuonH loss dynamics.} Experiments use Llama-124M trained on FineWeb. MuonW denotes Muon with weight decay, whereas MuonH uses Hyperball. In each subfigure, only the LR of the comparison run is adapted to match the target ELR, while its norm-control method remains unchanged. The compact panels show the LR and ELR trajectories, and the large panel compares the resulting losses. \textbf{(a)} MuonH serves as the target, and the LR of MuonW is adapted. \textbf{(b)} The matching direction is reversed: MuonW serves as the target, and the LR of MuonH is adapted. For MuonH, we use the optimizer-specific ELR defined in Appendix~\ref{appendix:exp_datails}.}
    \label{fig:app:act_muon_diagnostics}
\end{figure}

\paragraph{Hyperball with Muon.}
Figure~\ref{fig:app:act_muon_diagnostics}(a) supplements Figure~\ref{fig:wd_switch_lrnorm}(b) with the corresponding LR and ELR trajectories. MuonH, which uses Hyperball, serves as the target, while the LR of MuonW, which uses weight decay, is adapted to track its ELR schedule. Figure~\ref{fig:app:act_muon_diagnostics}(b) reverses the direction: MuonW serves as the target, and the LR of MuonH is adapted to track its ELR. The resulting loss trajectories remain closely aligned, with a mean collapse error of \(2.5\times10^{-3}\). Thus, the agreement is not specific to the direction of ELR matching.

\begin{figure}[!h]
    \centering
    \includegraphics[width=.9\textwidth]{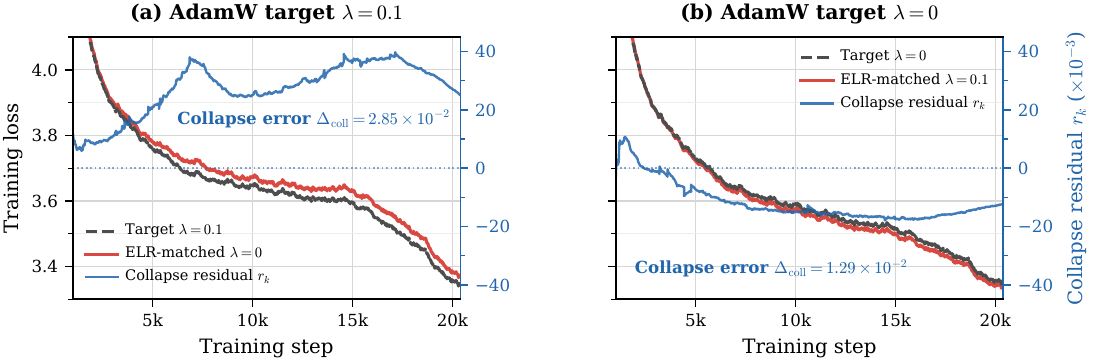}
\caption{\textbf{Fixing RMSNorm gains degrades LR-only ELR matching.} Dashed black curves show the target losses, red curves show the comparison runs whose LRs are adapted to match the target ELR schedules, and blue curves show the collapse residuals \(r_k\) on the right axis. \textbf{(a)} The run with \(\lambda=0.1\) serves as the target, and the LR of the \(\lambda=0\) run is adapted. \textbf{(b)} The matching direction is reversed. In both panels, all RMSNorm gains are fixed and no norm trajectory is prescribed. Compared with the learnable-gain results in Figure~\ref{fig:app:act_adamw_diagnostics}, fixing the gains produces sustained \(10^{-2}\)-scale residuals in both directions.}
    \label{fig:app:fixed_gain_diagnostics}
\end{figure}

\paragraph{Learnable RMSNorm gains remain important under LR-only intervention.}
Figure~\ref{fig:app:fixed_gain_diagnostics} repeats the AdamW LR-only intervention in Figure~\ref{fig:app:act_adamw_diagnostics}, with the RMSNorm gains fixed throughout training as the only change. As in the prescribed-ELR experiments, fixing these gains substantially weakens the collapse and increases the collapse error to the \(10^{-2}\) scale. Thus, the dependence on learnable gains is not specific to prescribed norm schedules; it persists when the parameter norm evolves freely and only the LR is adapted to match ELR.

\section{Functional Scaling Laws (Supplement to Section~\ref{sec:elr-fsl})}
\label{app:fsl-details}

We provide the FSL specification and per-trajectory results complementing Figure~\ref{fig:fsl-generalization} in Section~\ref{sec:elr-fsl}. 

\begin{figure}[!h]
    \centering
    \includegraphics[width=\linewidth]{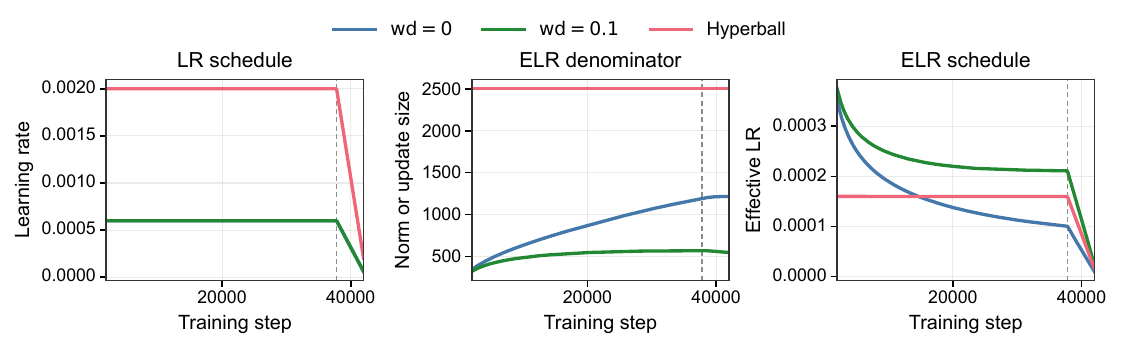}
    \caption{
        \textbf{LR schedule, ELR-denominator evolution, and ELR  schedule under distinct
        norm-control regimes.}
        All curves use WSD-10\%. Standard training uses weight decay with
        $\lambda \in\{0,0.1\}$. 
       For standard runs, the ELR denominator is the parameter Frobenius norm; for Hyperball, it is the average update norm over the post-warmup steps.
    }
    \label{fig:elr-fsl-rate-diagnostics}
\end{figure}

\paragraph{Trajectories and evaluation split.}
We train Llama-124M on FineWeb for 42,000 optimizer steps, including 1,000 warmup steps. Standard training uses $\mathrm{wd}\in\{0,0.1\}$ and a peak learning rate of $6\times10^{-4}$, whereas Hyperball directly controls the update norm and uses a peak learning rate of $2\times10^{-3}$. We consider four LR schedules. WSD-10\% and WSD-30\% decay LR over the final $10\%$ and $30\%$ of training, respectively. The 7111 schedule divides the post-warmup steps into four consecutive phases containing $70\%$, $10\%$, $10\%$, and $10\%$ of the steps, with LR reduced by a factor of $10^{1/3}$ at each phase boundary. The complete assignment of trajectories to the fitting, in-distribution (ID), and out-of-distribution (OOD) splits is reported in Table~\ref{tab:elr-fsl-per-curve}.  Figure~\ref{fig:elr-fsl-rate-diagnostics}
shows that ELR places the different norm-control regimes on a comparable
scale.

\paragraph{FSL parameterization and fitting.}
Given a rate schedule $\{r_j\}_{j\geq 0}$, we define the corresponding intrinsic time as
$
    t_k=\sum_{j=0}^{k} r_j.
$
FSL models the loss trajectory as
\begin{equation}
    \widehat L_k
    = L_\infty
    + C_1(B_1+t_k)^{-a_1}
    + C_2\sum_{j=0}^{k}
      (B_2+t_k-t_j)^{-a_2}r_j^2.
    \label{eq:appendix-fsl}
\end{equation}
The lr-FSL and elr-FSL models share this parameterization and differ only in the rate schedule used as input. Let $\eta_k$ denote the learning rate and $n_k$ the measured norm scale at step $k$. The two input schedules are
\begin{equation}
    r_k^{\mathrm{lr}}=\eta_k,
    \qquad
    r_k^{\mathrm{elr}}=\frac{\eta_k}{n_k}.
    \label{eq:appendix-rate-definitions}
\end{equation}
For standard training trajectories, $n_k$ is the Frobenius norm of the parameters; for Hyperball trajectories, it is the mean update-norm scale.

Following the FSL specification in the main text, we jointly fit the seven parameters
\[
    \Theta=(L_\infty,C_1,B_1,a_1,C_2,B_2,a_2)
\]
across the four fitting trajectories. For both variants, we smooth the loss trajectories using an exponential moving average with decay $0.99$, exclude the first $2{,}000$ steps from fitting, and use the same parameter initialization. We optimize $\Theta$ for 800 Adam steps, followed by 10 L-BFGS refinement steps.

\begin{figure}[t]
    \centering
    \includegraphics[width=\linewidth]{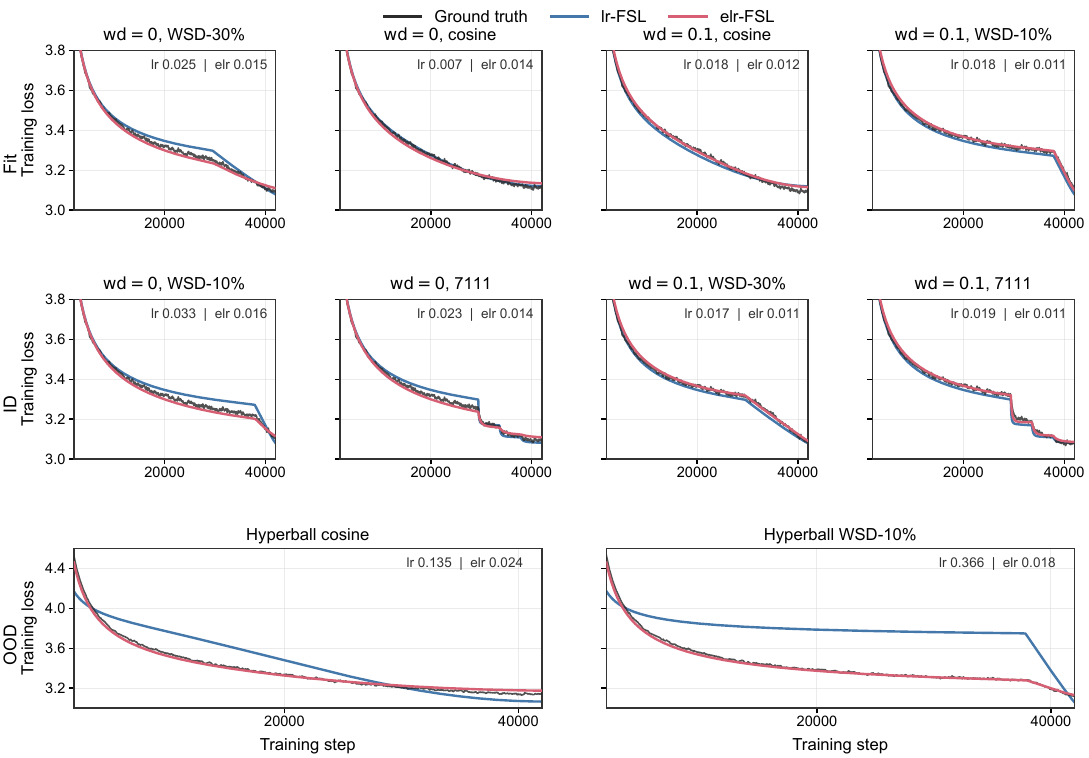}
    \caption{
        \textbf{Predictions for all fitted and held-out trajectories.}
        Rows show fit, ID schedule transfer, and OOD Hyperball trajectories.
        Black denotes ground truth; blue and red denote lr-FSL and elr-FSL.
        We see that elr-FSL is  substantially more accurate on both held-out ID  and OOD schedules.
    }
    \label{fig:elr-fsl-all-curves}
\end{figure}

\paragraph{Prediction Results.}
Table~\ref{tab:elr-fsl-per-curve} reports every trajectory, and
Figure~\ref{fig:elr-fsl-all-curves} shows the corresponding predictions.
elr-FSL has lower RMSE on nine of ten trajectories; under Hyperball, the
improvements reach $5.61\times$ for cosine and $19.96\times$ for WSD-10\%.

\begin{table}[ht]
    \centering
    \caption{
        \textbf{Per-trajectory prediction error.}
        Improvement = $\text{RMSE}_{lr}/\text{RMSE}_{elr}$; values above one
        favor elr-FSL. Bold marks the lower error.
    }
    \label{tab:elr-fsl-per-curve}
    \small
    \setlength{\tabcolsep}{8pt}
    \renewcommand{\arraystretch}{1.08}
    \begin{tabular}{@{}llccc@{}}
        \toprule
        Split & Trajectory
        & lr-FSL $\downarrow$
        & elr-FSL $\downarrow$
        & Improvement $\uparrow$ \\
        \midrule
        Fit & $\mathrm{wd}=0$, WSD-30\%   & 0.0251 & \textbf{0.0150} & $1.68\times$ \\
            & $\mathrm{wd}=0$, cosine     & \textbf{0.0069} & 0.0138 & $0.50\times$ \\
            & $\mathrm{wd}=0.1$, cosine   & 0.0180 & \textbf{0.0122} & $1.48\times$ \\
            & $\mathrm{wd}=0.1$, WSD-10\% & 0.0184 & \textbf{0.0112} & $1.65\times$ \\
        \midrule
        ID  & $\mathrm{wd}=0$, WSD-10\%   & 0.0327 & \textbf{0.0160} & $2.05\times$ \\
            & $\mathrm{wd}=0$, 7111       & 0.0233 & \textbf{0.0143} & $1.63\times$ \\
            & $\mathrm{wd}=0.1$, WSD-30\% & 0.0172 & \textbf{0.0110} & $1.56\times$ \\
            & $\mathrm{wd}=0.1$, 7111     & 0.0193 & \textbf{0.0113} & $1.70\times$ \\
        \midrule
        OOD & Hyperball, cosine   & 0.1353 & \textbf{0.0241} & $5.61\times$ \\
            & Hyperball, WSD-10\% & 0.3662 & \textbf{0.0183} & $19.96\times$ \\
        \bottomrule
    \end{tabular}
\end{table}

\FloatBarrier
\section{Delayed Acceleration (Supplement to Section~\ref{sec:delayed_acceleration})}
\label{appendix:delayed_acceleration}


\subsection{Acquired Early, Revealed Late: A Controlled WSD Test}
\label{appendix:wsd-delayed-acceleration}

The cosine-schedule experiments in the main text establish delayed acceleration. However, under cosine decay, the learning rate decreases continuously, making it difficult to vary the duration of the late low-ELR phase independently. We therefore conduct a complementary experiment using  WSD, whose terminal decay window can be controlled explicitly. Holding the peak LR and training budget fixed, we vary the WSD decay ratio from $0.1$ to $0.3$. This provides a targeted test of the temporal mechanism predicted by FSL.

FSL predicts that the benefit of weight decay is \emph{acquired early but revealed late}. By suppressing norm growth, weight decay sustains a larger ELR during the early phase. The resulting faster accumulation of effective training time improves the signal-learning term, but the larger ELR simultaneously produces greater noise accumulation. Consequently, the signal-learning advantage need not be visible immediately in the observed loss.
As ELR decays, new noise injection falls while previously accumulated noise continues to be forgotten, revealing the signal-learning advantage acquired earlier. The decisive prediction is therefore that a short decay phase may end before this advantage becomes visible, whereas a longer decay phase should reveal it and allow the weight-decayed run to overtake the unregularized baseline earlier.

\begin{figure}[!h]
    \centering
    \begin{minipage}[t]{0.90\textwidth}
        \centering
        \begin{minipage}[t]{0.48\linewidth}
            \centering
            \includegraphics[width=\linewidth]{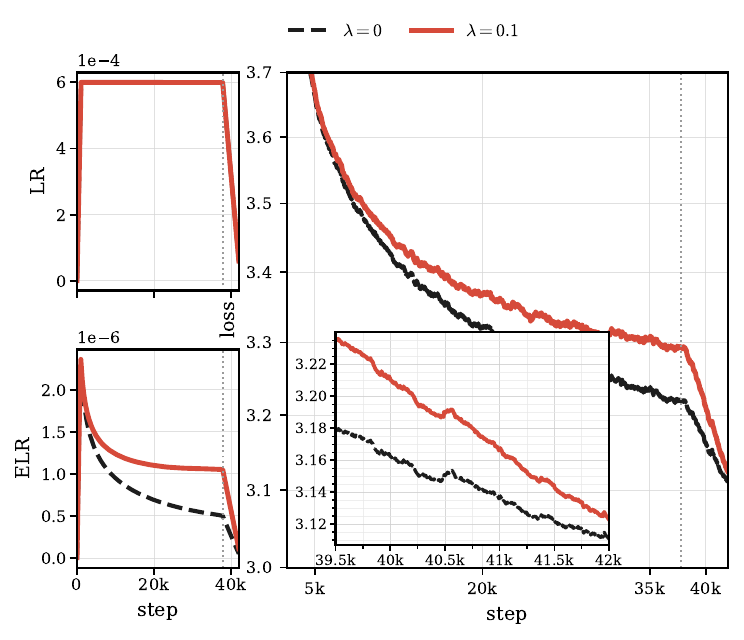}
            {\small\textbf{(a)} WSD decay ratio $0.1$}
        \end{minipage}
        \hspace{0.02\linewidth}
        \begin{minipage}[t]{0.48\linewidth}
            \centering
            \includegraphics[width=\linewidth]{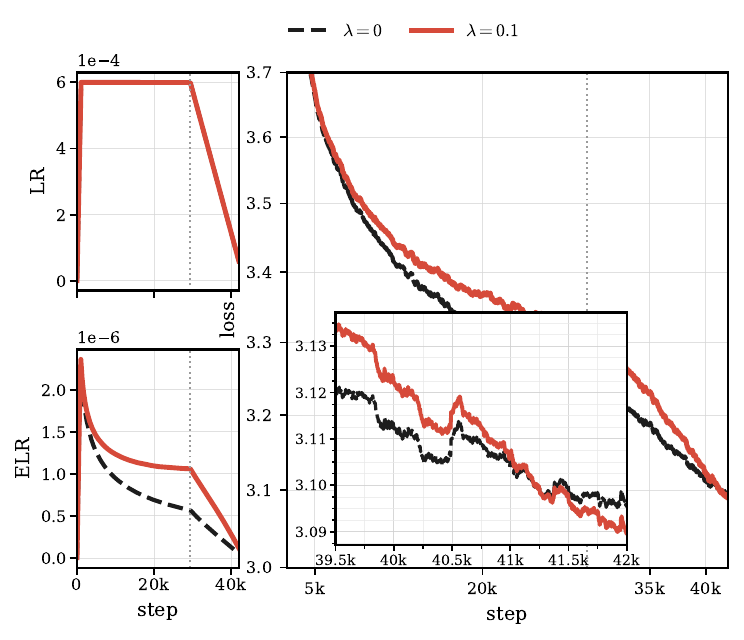}
            {\small\textbf{(b)} WSD decay ratio $0.3$}
        \end{minipage}
    \end{minipage}
\caption{\textbf{A longer WSD decay phase reveals the signal-learning gain acquired earlier.}
Using Llama-135M trained on FineWeb, we explicitly vary the duration of the terminal WSD decay phase. Each panel compares $\lambda\in\{0,0.1\}$ at a fixed WSD decay ratio; black and red denote $\lambda=0$ and $0.1$, respectively. With a decay ratio of $0.1$ (left), the decay phase is too short for the $\lambda=0.1$ run to overtake the unregularized baseline. Increasing the decay ratio to $0.3$ (right) extends the low-ELR phase, allowing the accumulated noise to be forgotten and revealing the earlier signal-learning advantage: the $\lambda=0.1$ run overtakes $\lambda=0$ and attains a lower final loss.}
\label{fig:wsd-decay-length}
\end{figure}

Figure~\ref{fig:wsd-decay-length} confirms this prediction. With a WSD decay ratio of $0.1$, the $\lambda=0.1$ run maintains a larger ELR than the $\lambda=0$ run, but the terminal decay phase is too short to reveal its signal-learning advantage: it remains above the unregularized baseline at the end of training. When the decay ratio is increased to $0.3$, the longer low-ELR phase changes the outcome. The $\lambda=0.1$ run overtakes $\lambda=0$ during decay and attains a lower final loss. In other words, the gain is accumulated before the crossover but becomes observable only during a sufficiently long decay phase. 

The predicted crossover under a longer decay window supports the FSL account: an early signal-learning gain is initially masked by noise and revealed only late.

\subsection{Delayed Acceleration under Hyperball}

\begin{figure}[!h]
    \centering
    \includegraphics[width=0.8\linewidth]{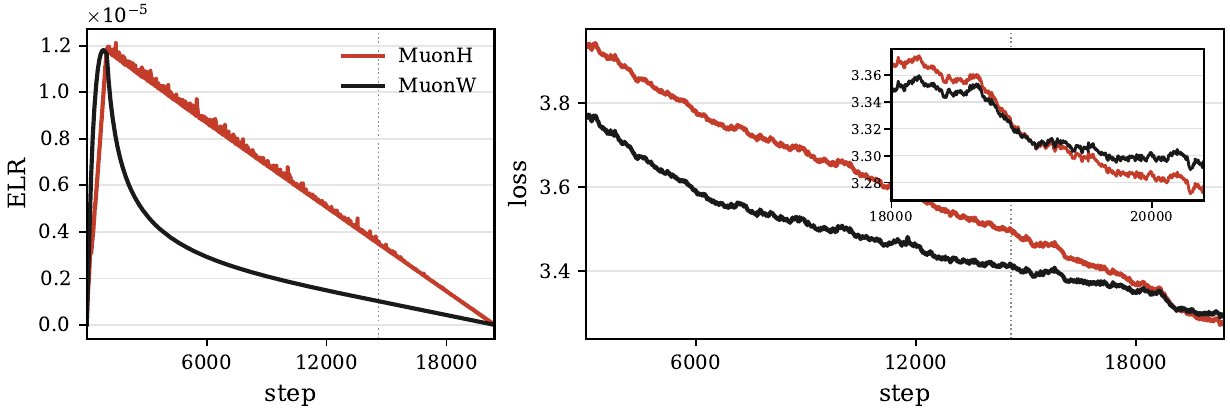}
    \caption{\textbf{Delayed acceleration under Hyperball.}
    On Llama-124M trained on FineWeb, MuonH initially has higher loss than MuonW, but later overtakes it and
    reaches a lower late-stage loss. Both runs use linear schedules.}
    \label{fig:app:hyperball_delayed_acceleration}
\end{figure}

The same qualitative phenomenon also appears under Hyperball. A delayed gain
is already visible in the MuonH trajectory reported by
\citet[Figure~3, right]{wen2026hyperball}. In our experiment,
Figure~\ref{fig:app:hyperball_delayed_acceleration} compares MuonH with MuonW
when both use linear decay. MuonH has higher loss
in early training, but subsequently catches up with and overtakes MuonW,
reaching a lower loss near the end of training. Delayed acceleration is
therefore not specific to weight decay; it also arises when the parameter norm
is controlled explicitly by Hyperball.

The same ELR mechanism explains this reversal: Hyperball's norm control
slows ELR decay relative to MuonW, maintaining a larger ELR before the
terminal decay.
 This accelerates signal learning but also increases noise
accumulation, initially masking the advantage.
During the terminal decay, the ELR decreases
and the accumulated noise is gradually forgotten, revealing the advantage
acquired earlier. This is consistent with
Figure~\ref{fig:wd_switch_lrnorm}(b), where matching the MuonH normalized
update coefficient reproduces its loss trajectory with a mean collapse error
of $1.2\times10^{-3}$.


\section{ELR Collapse in Vision Transformers}
\label{app:vit-elr-collapse}

\paragraph{Setup and ELR matching.}
To test whether ELR collapse extends beyond language modeling, we repeat the
norm-control matching experiment of Appendix~\ref{sub:elr_matching_protocols} on a
9.8M-parameter pre-norm  Vision Transformer (ViT)~\citep{dosovitskiy2020image} trained on ImageNet-1K~\citep{deng2009imagenet} with $64\times64$ inputs.
The model has 12 layers, hidden size 256, four attention heads, an MLP
expansion ratio of 4, and $8\times8$ patches; it uses RMSNorm, QK-Norm, and
ReLU activations. We train in FP32 for 50,000 steps using AdamW with  $(\beta_1,\beta_2)=(0.9,0.95)$, no weight decay, global batch size 512, and
a WSD schedule that warms up to $6\times10^{-4}$ over 1,250 steps, remains
constant until step 45,000, and decays to $6\times10^{-5}$.

We use the prescribed-ELR matching construction in
Appendix~\ref{sub:elr_matching_protocols}. For the ViT experiments, the controlled
parameter collections comprise the patch-projection affine parameters, the
class-token and positional embeddings, and the affine parameters in all
attention and MLP blocks. The
RMSNorm gains and classifier head retain their base learning rates and are
excluded from the prescribed-norm projection.

\begin{figure}[!h]
    \centering
    \includegraphics[width=0.92\linewidth]{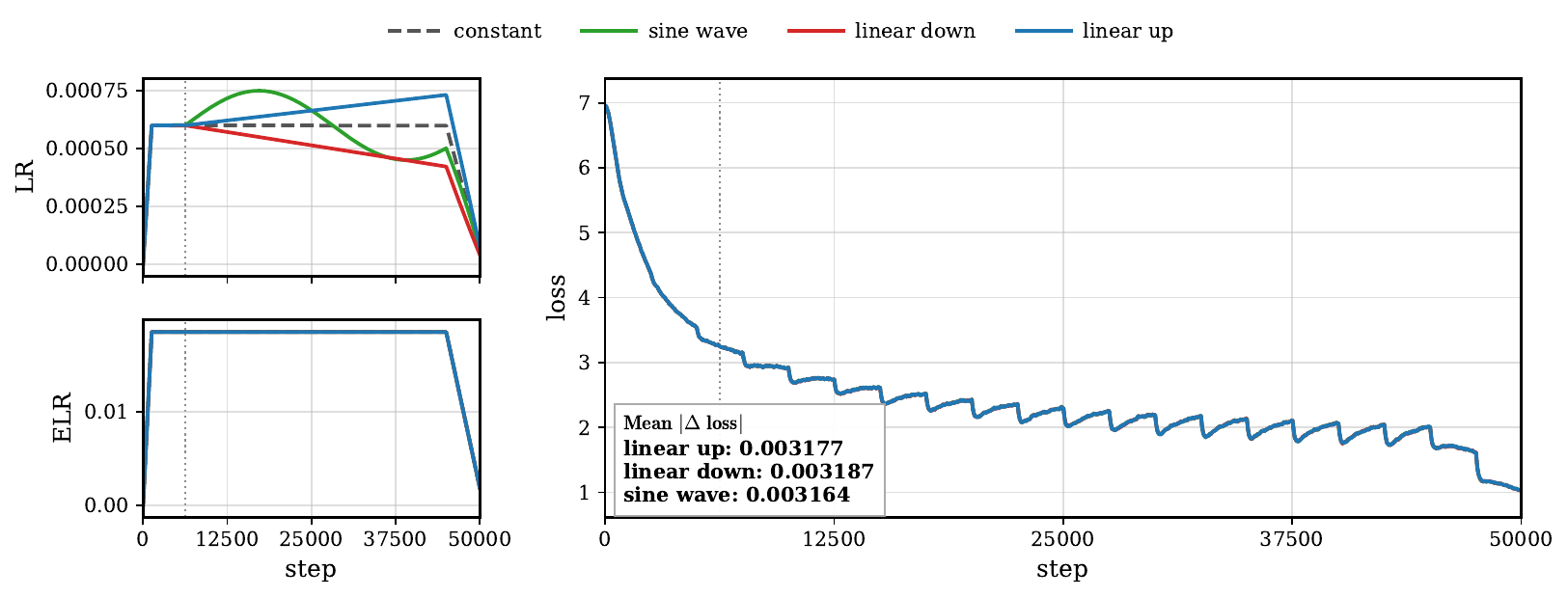}
\caption{\textbf{ELR collapse for a 9.8M-parameter ViT trained on ImageNet-1K.}
At step 6,250, training is branched into four runs with distinct LR and parameter-norm schedules but a matched ELR schedule. Their smoothed loss trajectories remain collapsed throughout the post-branch phase. The legend reports the collapse error $\Delta_{\collapse}$ of each run relative to the constant-LR reference.
}
    \label{fig:vit-elr-collapse}
\end{figure}

\paragraph{Result.}
We compute the mean collapse error as in the main text using the smoothed
training loss. Figure~\ref{fig:vit-elr-collapse} shows that the
three nonconstant branches remain aligned with the constant branch, with
$
\Delta_{\collapse}\in[3.16,3.19]\times10^{-3}.
$ 
This extends the $10^{-3}$-scale ELR
collapse observed in language model pretraining to this ViT setting.

\paragraph{Normalization also affects collapse precision in ViTs.}
We test whether the normalization dependence identified in Section~\ref{sec:conditions_for_elr_collapse} also appears in ViTs. Starting from the preceding configuration, we consider three nested variants: (i) QK-Norm with learnable gains, (ii) no QK-Norm with learnable gains, and (iii) no QK-Norm with all remaining RMSNorm gains fixed at initialization. For each variant, we repeat the prescribed-ELR protocol using the reference, linear-up, linear-down, and sinusoidal schedules.

\begin{figure}[h]
    \centering
    \includegraphics[width=0.9\linewidth]{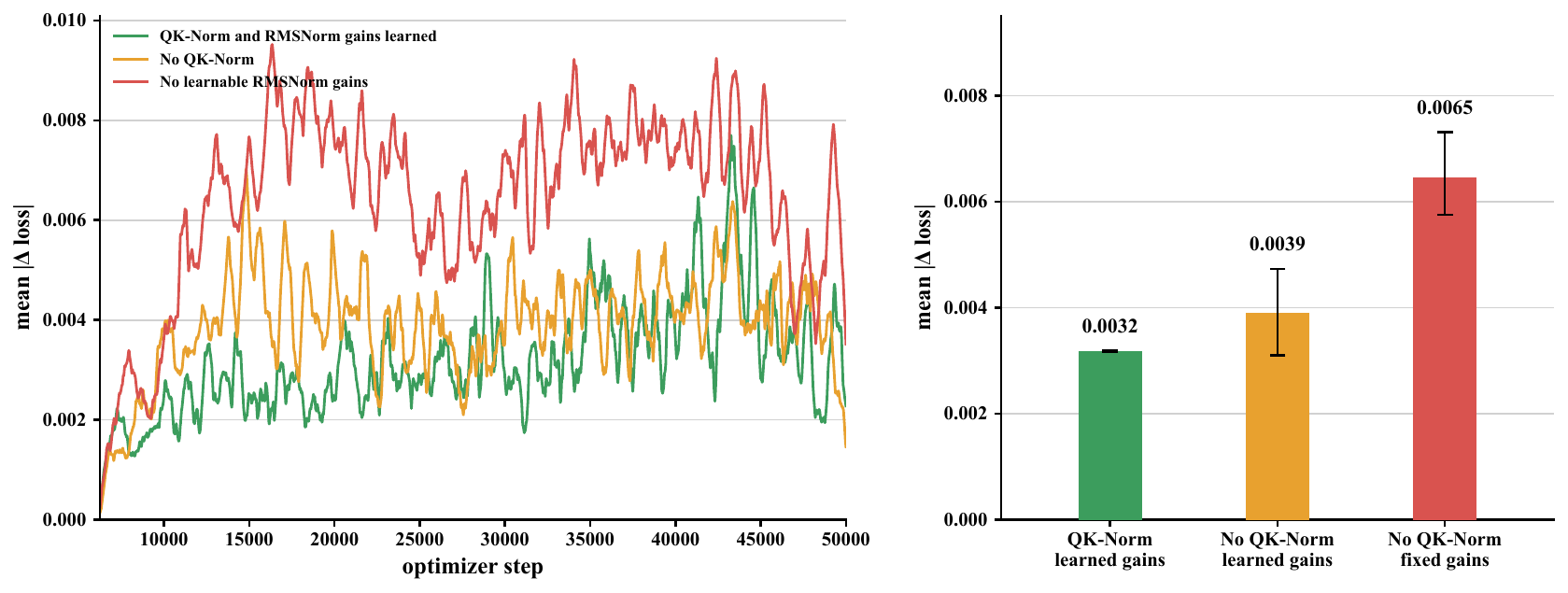}
\caption{\textbf{Normalization improves the precision of ELR collapse in
ViTs.}
\textbf{Left:} Mean absolute loss residual across the linear-up, linear-down,
and sine-wave ELR-matched schedules relative to the constant-LR reference
during training.
\textbf{Right:} Corresponding trajectory-averaged collapse error. Removing QK-Norm weakens
the collapse, while additionally fixing the remaining gains produces
the largest discrepancy.}
\label{fig:vit-ablation}
\end{figure}

Figure~\ref{fig:vit-ablation} shows the same qualitative ordering observed in language models. Averaged over the three nonreference schedules, the mean collapse error increases from \(3.18\times10^{-3}\) in the default configuration to \(3.90\times10^{-3}\) after removing QK-Norm, and further to \(6.46\times10^{-3}\) after fixing the RMSNorm gains. This result shows that the normalization dependence of ELR collapse is not confined to our language model experiments, but also appears in the ViT/ImageNet setting studied here.

\end{document}